\documentclass[manuscript]{acmart}

\AtBeginDocument{%
	\providecommand\BibTeX{{%
			\normalfont B\kern-0.5em{\scshape i\kern-0.25em b}\kern-0.8em\TeX}}}

\usepackage[]{algorithm2e}
\usepackage{float}
\RestyleAlgo{boxruled}

\newcommand{\varSymbol}[3]{
\ifx \\#2\\
	\ifx \\#3\\ \lowercase{#1}
	\else
	\lowercase{#1}^{#3}
	\fi
\else
	\ifx \\#3\\ \lowercase{#1}_{#2}
	\else
		\lowercase{#1}_{#2}^{#3}
	\fi
\fi
}

\newcommand{\optimalVarSymbol}[3]
{
	\ifx\\#2\\
	\ifx\\#3\\ \optimal{\lowercase{#1}}
	\else
	\optimal{\lowercase{#1}}\textsuperscript{#3}
	\fi
	\else
	\ifx\\#3\\ \optimal{\lowercase{#1}}\textsubscript{#2}
	\else
	\optimal{\lowercase{#1}}\textsubscript{#2}\textsuperscript{#3}
	\fi
	\fi
}

\newcommand{\specialSetSymbol}[3]
{
	\ifx\\#2\\ \ifx\\#3\\  \overline{\uppercase{#1}}
	\else
	 \overline{\uppercase{#1}}\textsuperscript{#3}
	\fi
	\else
	\ifx\\#3\\  \overline{\uppercase{#1}}\textsubscript{#2}
	\else
	 \overline{\uppercase{#1}}\textsubscript{#2}\textsuperscript{#3}
	\fi
	\fi
}

\newcommand{\optimalSetSymbol}[3]
{
	\ifx\\#2\\ \ifx\\#3\\ \optimal{\uppercase{#1}}
	\else
	\optimal{\uppercase{#1}}\textsuperscript{#3}
	\fi
	\else
	\ifx\\#3\\ \optimal{\uppercase{#1}}\textsubscript{#2}
	\else
	\optimal{\uppercase{#1}}\textsubscript{#2}\textsuperscript{#3}
	\fi
	\fi
}

\newcommand{\estimatedOptimalVarSymbol}[3]
{
	\ifx\\#2\\
	\ifx\\#3\\ \optimal{\widetilde{\lowercase{#1}}}
	\else
	\optimal{\widetilde{\lowercase{#1}}}\textsuperscript{#3}
	\fi
	\else
	\ifx\\#3\\ \optimal{\widetilde{\lowercase{#1}}}\textsubscript{#2}
	\else
	\optimal{\widetilde{\lowercase{#1}}}\textsubscript{#2}\textsuperscript{#3}
	\fi
	\fi
}

\newcommand{\estimatedOptimalSetSymbol}[3]
{
	\ifx\\#2\\
	\ifx\\#3\\ \optimal{\widetilde{\uppercase{#1}}}
	\else \optimal{\widetilde{\uppercase{#1}}}\textsuperscript{#3}
	\fi
	\else \ifx\\#3\\ \optimal{\widetilde{\uppercase{#1}}}\textsubscript{#2}
	\else \optimal{\widetilde{\uppercase{#1}}}\textsubscript{#2}\textsuperscript{#3}
	\fi
	\fi
}

\newcommand{\functionSymbol}[1]{#1}
\newcommand{\functionTextSymbol}[1]{\texttt{#1}}

\newcommand{\functionFormalSignature}[2]{
	\functionSymbol{#1}\colon #2
}

\newcommand{\functionFormal}[3]{
	\functionFormalSignature{#1}{#2} \rightarrow #3
}

\newcommand{\functionSignature}[2]{
\ifx \\#2\\
	\functionSymbol{#1}
\else
	\functionSymbol{#1}(#2)
\fi
}

\newcommand{\setSymbol}[3] {
\ifx\\#2\\\ifx\\#3\\\uppercase{#1}
\else
\uppercase{#1}^{#3}
\fi
\else
\ifx\\#3\\\uppercase{#1}_{#2}
\else
\uppercase{#1}_{#2}^{#3}
\fi
\fi
}

\newcommand{\varSymbolHat}[3]
{
	\ifx \\#2\\
	\ifx \\#3\\
	\widehat{\lowercase{#1}}
	\else
	\widehat{\lowercase{#1}}\textsuperscript{#3}
	\fi
	\else
	\ifx \\#3\\
	\widehat{\lowercase{#1}}\textsubscript{#2}
	\else
	\widehat{\lowercase{#1}}\textsubscript{#2}\textsuperscript{#3}
	\fi
	\fi
}
\newcommand{\setSymbolHat}[3]
{
	\ifx\\#2\\
	\ifx\\#3\\
	\widehat{\uppercase{#1}}
	\else
	\widehat{\uppercase{#1}}\textsuperscript{#3}
	\fi
	\else
	\ifx\\#3\\
	\widehat{\uppercase{#1}}\textsubscript{#2}
	\else
	\widehat{\uppercase{#1}}\textsubscript{#2}\textsuperscript{#3}
	\fi
	\fi
}
\newcommand{\varSymbolPlus}[2]
{
	\ifx\\#2\\
	{\lowercase{#1}}\textsuperscript{$\oplus$}
	\else
	{\lowercase{#1}}\textsubscript{#2}\textsuperscript{\oplus}
	\fi
}

\newcommand{\varSymbolMinus}[2]
{
	\ifx\\#2\\
	{\lowercase{#1}}\textsuperscript{$\ominus$}
	\else
	{\lowercase{#1}}\textsubscript{#2}\textsuperscript{$\ominus$}
	\fi
}

\newcommand{\setSymbolPlus}[2]
{
	\ifx\\#2\\
	{\uppercase{#1}}\textsuperscript{$\oplus$}
	\else
	{\uppercase{#1}}\textsubscript{#2}\textsuperscript{\oplus}
	\fi
}

\newcommand{\setSymbolMinus}[2]
{
	\ifx\\#2\\
	{\uppercase{#1}}\textsuperscript{$\ominus$}
	\else
	{\uppercase{#1}}\textsubscript{#2}\textsuperscript{$\ominus$}
	\fi
}

\newcommand{\setOfSetsSymbolMinus}[2]
{
	\ifx\\#2\\
	{\mathcal{\uppercase{#1}}}\textsuperscript{$\ominus$}
	\else
	{\mathcal{\uppercase{#1}}}\textsubscript{#2}\textsuperscript{$\ominus$}
	\fi
}

\newcommand{\varSymbolHatPlus}[2]
{
	\ifx\\#2\\
	\widehat{\lowercase{#1}}\textsuperscript{$\oplus$}
	\else
	\widehat{\lowercase{#1}}\textsubscript{#2}\textsuperscript{\oplus}
	\fi
}

\newcommand{\setSymbolHatPlus}[2]
{
	\ifx\\#2\\
	\widehat{\uppercase{#1}}\textsuperscript{$\oplus$}
	\else
	\widehat{\uppercase{#1}}\textsubscript{#2}\textsuperscript{\oplus}
	\fi
}

\newcommand{\varSymbolHatMinus}[2]
{
	\ifx\\#2\\
	\widehat{\lowercase{#1}}\textsuperscript{$\ominus$}
	\else
	\widehat{\lowercase{#1}}\textsubscript{#2}\textsuperscript{$\ominus$}
	\fi
}

\newcommand{\setSymbolHatMinus}[2]
{
	\ifx\\#2\\
	\widehat{\uppercase{#1}}\textsuperscript{$\ominus$}
	\else
	\widehat{\uppercase{#1}}\textsubscript{#2}\textsuperscript{$\ominus$}
	\fi
}

\newcommand{\setOfSetsSymbolHatMinus}[2]
{
	\ifx\\#2\\
	\widehat{\mathcal{\uppercase{#1}}}\textsuperscript{$\ominus$}
	\else
	\widehat{\mathcal{\uppercase{#1}}}\textsubscript{#2}\textsuperscript{$\ominus$}
	\fi
}

\newcommand{\setOptimalSymbol}[3]
{
	\ifx\\#2\\
	\ifx\\#3\\
	\optimal{\uppercase{#1}}
	\else
	\optimal{\uppercase{#1}}\textsuperscript{#3}
	\fi
	\else
	\ifx\\#3\\
	\optimal{\uppercase{#1}}\textsubscript{#2}
	\else
	\optimal{\uppercase{#1}}\textsubscript{#2}\textsuperscript{#3}
	\fi
	\fi
}

\newcommand{\powerSetSymbolP}[3]
{
	\ifx \\#2\\ \ifx\\#3\\ \mathcal{P}(\uppercase{#1})
	\else \mathcal{P}({\uppercase{#1}})\textsuperscript{#3}
	\fi
	\else
	\ifx\\#3\\ \mathcal{P}({\uppercase{#1}})\textsubscript{#2}
	\else
	\mathcal{P}({\uppercase{#1}})\textsubscript{#2}\textsuperscript{#3}
	\fi
	\fi
}

\newcommand{\setOfSetsSymbol}[3]{\ifx\\#2\\\ifx\\#3\\{\mathcal{\uppercase{#1}}}\else {\mathcal{\uppercase{#1}}}\textsuperscript{#3}\fi\else \ifx\\#3\\{\mathcal{\uppercase{#1}}}\textsubscript{#2}\else{\mathcal{\uppercase{#1}}}\textsubscript{#2}\textsuperscript{#3}\fi\fi}

\newcommand{\setOfSetsOptimalSymbol}[3]
{
	\ifx\\#2\\ \ifx\\#3\\ \optimal{\mathcal{\uppercase{#1}}}
	\else
	\optimal{\mathcal{\uppercase{#1}}}\textsuperscript{#3}
	\fi
	\else
	\ifx\\#3\\ \optimal{\mathcal{\uppercase{#1}}}\textsubscript{#2}
	\else
	\optimal{\mathcal{\uppercase{#1}}}\textsubscript{#2}\textsuperscript{#3}
	\fi
	\fi
}

\newcommand{\optimal}[1]{\accentset{\ast}{#1}}

\newcommand{\setEstimatedSymbol}[3]
{
	\ifx\\#2\\
	\ifx\\#3\\
	\estimated{\uppercase{#1}}
	\else
	\estimated{\uppercase{#1}}\textsuperscript{#3}
	\fi
	\else
	\ifx\\#3\\
	\estimated{\uppercase{#1}}\textsubscript{#2}
	\else
	\estimated{\uppercase{#1}}\textsubscript{#2}\textsuperscript{#3}
	\fi
	\fi
}

\newcommand{\estimated}[1]
{
	\widehat{{#1}}
}

\newcommand{\matrixSymbol}[3] {
	\ifx\\#2\\\ifx\\#3\\\textbf{\uppercase{#1}}
	\else
	\uppercase{\textbf{#1}}^{#3}
	\fi
	\else
	\ifx\\#3\\\uppercase{\textbf{#1}}_{#2}
	\else
	\uppercase{\textbf{#1}}_{#2}^{#3}
	\fi
	\fi
}

\newcommand{\scriptSymbol}[3]{\ifx\\#2\\
	\mathcal{\uppercase{#1}}
	\else
	\ifx\\#3\\
	\mathcal{\uppercase{#1}}\textsubscript{#2}
	\else
	\mathcal{\uppercase{#1}}\textsubscript{#2}\textsuperscript{#3}
	\fi
	\fi
}

\def\capitaliseaux#1#2\relax{\uppercase{#1}\lowercase{#2}}

\newcommand{\setBuilder}[3]{
	#1
	\IfSubStr{#1}{(}{\funcdef}{=}
	\lbrace
	\IfSubStr{#2}{,}{(#2)}{#2}
	\suchthat
	#3
	\rbrace
}

\newcommand{\funcdef}{=}

\newcommand{\funcupdate}{\leftarrow}
\newcommand{\suchthat}{:\ }

\newcommand{\setRealNumbersNonNegative}[2]{\mathbb{R}_{>=0}^{#2}}

\newcommand{\setIntegersPositive}[2]{\mathbb{Z}{+}{#2}}
\newcommand{\setIntegersNonNegative}[2]{\mathbb{Z}_{>=0}^{#2}}

\newcommand{\varX}[2]{\varSymbol{x}{#1}{#2}}

\newcommand{\setX}[2]{\setSymbol{x}{#1}{#2}}

\newcommand{\funcSumNorm}[2]{
	\ifx\\#2\\ \functionTextSymbol{sumnorm}(#1)
	\else
	\functionTextSymbol{sumnorm}_{ax=#2}(#1)
	\fi
}

\newcommand{\functionBoltzmann}[2]{
\ifx \\#1\\
	p_i = \dfrac{e^{(q_i/\tau)}}{\sum_{j=1}^{N} e^{(q_j/\tau)}}
\else
p_{#1} = \dfrac{e^{(q_{#1}/\tau)}}{\sum_{j=1}^{N} e^{(q_j/\tau)}}
\fi
}

\newcommand{\funcSize}[1]{\lvert #1 \rvert}

\newcommand{\equationDefinition}[3]{
\textbf{#1}
\ifx \\#2\\ \else \textit{#2} \fi
\ifx \\#3\\ \else
\begin{equation}
#3
\end{equation}
 \fi
}

\newcommand{\setUniform}[2]{\setSymbol{U}{#1}{#2}}

\newcommand{\acronymResourceAllocationAlgorithm}[2]{MG-RAO}
\newcommand{\acronymResourceAllocationAlgorithmExtended}[2]{multi-group resource allocation optimisation (MG-RAO)}

\newcommand{\simulationUniform}[2]{<\texttt{uniform}>}
\newcommand{\simulationMgrao}[2]{<\texttt{mgrao}>}
\newcommand{\simulationMulti}[2]{<\texttt{multi}>}
\newcommand{\simulationMgraoOne}[2]{<\texttt{mgrao-}1:1>}
\newcommand{\simulationMgraoXY}[2]{<\texttt{mgrao-}x:y>}
\newcommand{\simulationMgraoMicro}[2]{<\texttt{mgrao-}2:1>}
\newcommand{\simulationMgraoTiny}[2]{<\texttt{mgrao-}5:1>}
\newcommand{\simulationMgraoSmall}[2]{<\texttt{mgrao-}10:1>}
\newcommand{\simulationMgraoMedium}[2]{<\texttt{mgrao-}25:1>}
\newcommand{\simulationMgraoLarge}[2]{<\texttt{mgrao-}50:1>}
\newcommand{\simulationMgraoMax}[2]{<\texttt{mgrao-max}>}

\newcommand{\resultMgraoMaxToMax}[2]{0.0\%}
\newcommand{\resultMgraoMaxPercentageMax}[2]{100.0\%}

\newcommand{\resultSingleMgraoMax}[2]{28.0\%}
\newcommand{\resultSingleMgraoOneOne}[2]{23.4\%}
\newcommand{\resultSingleMgraoMaxToOneOne}[2]{4.6\%}
\newcommand{\resultSingleMgraoOneOnePercentageMax}[2]{83.7\%}

\newcommand{\resultMultiMgraoMax}[2]{23.8\%}
\newcommand{\resultMultiMgraoOneOne}[2]{21.4\%}
\newcommand{\resultMultiMgraoMaxToOneOne}[2]{2.4\%}
\newcommand{\resultMultiMgraoOneOnePercentageMax}[2]{90.0\%}

\newcommand{\resultLargeMgraoMax}[2]{19.6\%}

\newcommand{\resultLargeMgraoOneOne}[2]{14.7\%}
\newcommand{\resultLargeMgraoMaxToOneOne}[2]{4.9\%}
\newcommand{\resultLargeMgraoOneOnePercentageMax}[2]{75.1\%}

\newcommand{\resultLargeMgraoTwoOne}[2]{17.2\%}
\newcommand{\resultLargeMgraoMaxToTwoOne}[2]{2.4\%}
\newcommand{\resultLargeMgraoTwoOnePercentageMax}[2]{87.7\%}

\newcommand{\resultLargeMgraoFiveOne}[2]{17.9\%}
\newcommand{\resultLargeMgraoMaxToFiveOne}[2]{1.8\%}
\newcommand{\resultLargeMgraoFiveOnePercentageMax}[2]{91.0\%}

\newcommand{\resultLargeMgraoTenOne}[2]{18.9\%}
\newcommand{\resultLargeMgraoMaxToTenOne}[2]{0.6\%}
\newcommand{\resultLargeMgraoTenOnePercentageMax}[2]{96.6\%}

\newcommand{\resultLargeMgraoTwentyFiveOne}[2]{19.5\%}
\newcommand{\resultLargeMgraoMaxToTwentyFiveOne}[2]{0.1\%}
\newcommand{\resultLargeMgraoTwentyFiveOnePercentageMax}[2]{99.4\%}

\newcommand{\resultVolatileMgraoMax}[2]{7.1\%}
\newcommand{\resultVolatileMgraoOneOne}[2]{3.3\%}
\newcommand{\resultVolatileMgraoMaxToOneOne}[2]{3.8\%}
\newcommand{\resultVolatileMgraoOneOnePercentageMax}[2]{46.5\%}

\newcommand{\resultRangeFixedMgrao}[2]{23-28\%}
\newcommand{\resultRangeDirectMgrao}[2]{3-7\%}

\acmJournal{TAAS}
\acmVolume{0}
\acmNumber{0}
\acmArticle{0}
\acmYear{2021}
\acmMonth{0}
\setcopyright{acmlicensed}
\copyrightyear{2021}

\acmDOI{0000001.0000001}

\begin{document}
\title{Resource allocation in dynamic multiagent systems}
	
\author{Niall Creech}
\affiliation{%
	\institution{Kings College London}
	\department{Department of Informatics}
	\city{London}
	\postcode{WC2B 4BG}
	\country{UK}
}
\email{niall.creech@kcl.ac.uk}

\author{Natalia Criado Pacheco}
\affiliation{%
	\institution{Kings College London}
	\department{Department of Informatics}
	\city{London}
	\postcode{WC2B 4BG}
	\country{UK}
}
\email{natalia.criado_pacheco@kcl.ac.uk}

\author{Simon Miles}
\affiliation{%
	\institution{Kings College London}
	\department{Department of Informatics}
	\city{London}
	\postcode{WC2B 4BG}
	\country{UK}
}
\email{simon.miles@kcl.ac.uk}

\begin{abstract}
Resource allocation and task prioritisation are key problem domains in the fields of autonomous vehicles, networking, and cloud computing. The challenge in developing efficient and robust algorithms comes from the dynamic nature of these systems, with many components communicating and interacting in complex ways. The \acronymResourceAllocationAlgorithmExtended{}{} algorithm we present uses multiple function approximations of resource demand over time, alongside reinforcement learning techniques, to develop a novel method of optimising resource allocation in these multi-agent systems. This method is applicable where there are competing demands for shared resources, or in task prioritisation problems.

Evaluation is carried out in a simulated environment containing multiple
competing agents. We compare the new algorithm to an approach where child
agents distribute their resources uniformly across all the tasks they can be
allocated. We also contrast the performance of the algorithm where resource
allocation is modelled separately for groups of agents, as to being modelled jointly over all agents. The \acronymResourceAllocationAlgorithm{}{} algorithm shows a $\resultRangeFixedMgrao{}{}$ improvement over fixed resource allocation in the simulated environments. Results also show that, in a volatile system, using the \acronymResourceAllocationAlgorithm{}{} algorithm configured so that child agents model resource allocation for all agents as a whole has $\resultVolatileMgraoOneOnePercentageMax{}{}$ of the performance of when it is set to model multiple groups of agents. These results demonstrate the ability of the algorithm to solve resource allocation problems in multi-agent systems and to perform well in dynamic environments.
\end{abstract}

\ccsdesc[500]{Computing methodologies~Multi-agent systems}
\ccsdesc[500]{Theory of computation~Multi-agent reinforcement learning}
\ccsdesc[500]{Computing methodologies~Intelligent agents}
\ccsdesc[300]{Computing methodologies~Multi-agent planning}
\ccsdesc[300]{Theory of computation~Multi-agent learning}
\ccsdesc[300]{Computing methodologies~Mobile agents}
\ccsdesc[300]{Computing methodologies~Cooperation and coordination}
\ccsdesc[300]{Computing methodologies~Q-learning}
	
\keywords{Multi-agent systems, Distributed task allocation, Distributed Q-learning, Multi-agent reinforcement learning, MARL, MANET, V2X, VANET}
	
\renewcommand{\shortauthors}{N. Creech et al.}
	
\maketitle
\section{Introduction}
\label{section:introduction}
Prioritising tasks and resource allocation in distributed systems is an increasing challenge in real-world problems. As such, there is practical value in developing algorithms that can operate well in these environments, in particular, in agent-based environments where there is a large amount of interaction, uncertainty, or dynamism. There are many problem areas where scalability challenges limit the effectiveness of centralised techniques for orchestration, or where the poor reliability of communications requires some level of autonomous behaviour. Significant research on these problems has been carried out in areas such as,

\begin{itemize}
	\item Vehicular ad-hoc networks (VANET) and traffic control \cite{Althamary2019,Tong2019,Arel2010}
	\item Unmanned autonomous vehicles (UAV) communication and power management \cite{Cui2020}
	\item Routing and power management in wireless sensor networks \cite{Jafarzadeh2014}
	\item Resource allocation and service scaling in cloud computing \cite{Dutreilh2011}
	\item  Quality of service in content delivery networks \cite{Yu2008a}.
\end{itemize}
Current solutions often target specific systems and technologies but share commonalities due to the underlying theoretical problem being similar. These can be broadly categorised into centralised or decentralised solutions. In centralised solutions the behaviour of agents is coordinated through a shared decision-making component. As the environments become more dynamic and the number of interacting agents expands, the complexity of orchestration and communication increases \cite{Chevaleyre2005,Briola2011,Chevaleyre}. The use of hierarchical structures, such as in \textit{holonic systems}, can increase the applicability of this approach but limits in scalability still exist due to the same issues \cite{Fischer2003,Rodriguez2011,Wang2020a}. 

With decentralised approaches, agents behave with at least some autonomy. They have a local-view of the system, not a global one, either due to the system being partially observable to them or too complex for them to use system-wide knowledge effectively. Solutions that utilise system-wide information can optimise well, however do not scale well as the complexity of the systems increase and calculations become intractable. When local-knowledge only us used, scalability is increased, but optimisation can be difficult due to the use of multiple independent solutions based on agents' local views instead of a combined global solution.

The solution we present here aims to solve the problem of resource allocation and task prioritisation in distributed multi-agent systems where scalability and system dynamism are important factors.  The \textit{\acronymResourceAllocationAlgorithmExtended{}{}} algorithm combines update and prioritisation sub-algorithms, which are designed to overcome the issues of the other approaches described above. Using this Q-learning based algorithm an agent learns a group of function approximations mapping its incoming resource demands to its existing resources. This allows it to adapt its allocation policy and prioritise incoming tasks optimally. The algorithm responds to dynamic environments where agents join and leave, resource demands vary, and where the distribution pattern of requests may change over time.

The rest of the paper will introduce the relevant background in Section
\ref{section:background} before describing the problem explicitly in Sections
\ref{section:problem}, followed by the detail of our solution and algorithms in Section \ref{section:solution}. We evaluate the performance of our algorithms through simulation and analysis in Section \ref{section:analysis}. Finally we look at the conclusions drawn and future research in Section \ref{section:conclusions}.

\section{Related work}
\label{section:background}

Many different industries involve multi-agent systems where there are
\textit{resource allocation} problems, with multiple agents' placing demands on
shared resources, or \textit{task-prioritisation} problems, where multiple
agents' task requests to other agents compete for prioritisation. There are close-similarities between these problem types in-terms of the algorithms that can be applied. In addition, the similarity of the underlying systems across multiple industries makes research in one area highly relevant to many others. As such, although we focus on applications and research within multi-vehicle systems, many of the insights transfer to other problem domains and industry sectors mentioned previously.

Modelling and coordinating behaviours of actors in
\textit{vehicle-to-everything (V2X) systems} is essential for autonomous
driving and interconnected transport management. Information exchange allows
for vehicles to be more aware of other vehicles \cite{Nothdurft2011}, and to
interact with road infrastructure systems. This means they can make decisions
with increased safety and greater efficiency \cite{Xie2017a}. A request between
vehicles may be a composite set of tasks such as providing position and speed
data, traffic congestion information \cite{Schunemann2010}, traffic light
status, road-blockages, and so on. Each of these tasks requires the vehicle
serving the request to dedicate resources to acquiring and aggregating
information. Each of these tasks also has a varying degree of value to the
vehicle receiving the data depending on its situation. A vehicle travelling at
speed may prefer nearby vehicles to provide position and velocity data. When
vehicles are further away then less immediate factors such as traffic density
ahead may become more valuable \cite{Rizzo2016,Eiza2015}. A vehicle may receive
such tasks from many vehicles and must decide how to prioritise and allocate the required resources amongst these competing demands. The interactions can also be highly dynamic. Nearby vehicles may communicate updates at a high frequency, becoming less frequent as they move further away. Interruptions to connectivity can come from buildings, other vehicles, and interference. \cite{BelagalMath2017,Wang2019a}. With each vehicle providing its own set of resources, there is also the possibility of forming an ad-hoc distributed compute platform amongst multiple vehicles, which requires efficient resource allocation to handle the many distributed tasks in the system \cite{Feng2017,Xu2020}. Similar problems arise in UAV and other MANET (Mobile ad-hoc networks) more generally. There may be multiple vehicle-to-vehicle and vehicle-to-ground communications, and restricted energy availability, which needs to be managed effectively \cite{Cui}. The interactions of multiple mobile vehicles is an area where there is a large body of research on artificial intelligence applications \cite{Althamary2019,Tong2019} to develop models and predictive solutions.

V2X systems highlight the key elements of the systems we look to provide a solution for,
\begin{enumerate}
	 \item {
	 	a frequency of incoming composite tasks which require decomposition and prioritisation}
	 \item{
	 	the value of each task is dependant on the unknown state of the requester
	 	
	 }
	\item {
		there are multiple competing tasks demanding resources
	}
	\item {
 	these elements vary over time in both value and reliability.
	}
\end{enumerate}

Systems such as these are categorisable as \textit{multi-agent resource allocation} \cite{Briola2011} or \textit{dynamic task-scheduling} problems \cite{Shyalika2020}. For these there are established methods that can be applied such as auction protocols and automated negotiation schemes \cite{Beam1997} \cite{Fatima2001, Chevaleyre2005}, e.g. contract-net \cite{Smith1980a} and its more recent extensions \cite{Bozdag2008}. In these, an agent announces the availability of some resource it owns, agents interested in gaining access to the resource make bids, then the resource owner makes the final allocation. Although broadly used in many areas, the limitations become apparent when applied to large-scale and complex systems. 

The first problem occurs due to the \textit{lack of scalability with increased number of participating agents}. When there are many agents bidding or allocating resources, the negotiation process carries a significant overhead. This can delay, or otherwise reduce the optimality of, the overall allocation. Extensions such as concurrent contract-net \cite{Chevaleyre} mitigate some of those effects by extending the negotiation protocol but inherently have the same limitations. 

The second issue comes from the \textit{complexity of resource allocation effects}. The allocation of resources to one agent can positively or negatively impact further agents whose own demands rely on that agents' ability to acquire those resources. In addition, where there are many agents requesting a resource-type, and many that can allocate that resource, finding the optimal distribution of those agents' requests across those resource-allocating agents is a challenging problem. Here \textit{combinatorial auctions} \cite{Vriesde1998,Parkes2000} can help develop more complex allocation strategies in these situations, but with an associated negative impact on the scalability of solutions.   
 
The longer-term effects of resource allocations can also be modelled through the use of \textit{multi-agent reinforcement learning} \cite{Busoniu2008a} to learn and adapt allocation policies. This helps learn more complex relationships between resource allocation actions. In doing so we can use two main strategies. \textit{Joint action learning} uses algorithms based on Q-learning or deep learning to learn a model for the system as a whole, using all agents' knowledge in the system combined. In \textit{independent action learning}, each agent learns independently of other agents in the system using only a localised view of the system \cite{Claus}, without coordination. Distributed Q-learning algorithms provide some of the state-of-the-art examples in this area  \cite{Lauer00analgorithm, 4399095}.

Whereas we can best solve the problem of global resource allocation optimisation through joint-action learning, this becomes computationally intractable as the number of agents and states increases, so suffers from the scalability problem. Alternatively, independent action learning avoids the costs of intercommunication with other agents so is more scalable, but without a system-wide view does not optimise as well. This is due to the lack of observability of other agents' strategies, and risks becoming stuck in locally-optimal solutions. \cite{Fatima2001,Hernandez-Leal2019a, Shyalika2020}. 

Our approach is to extend the localised view by enabling a \textit{parent agent}, the resource-requesting agent, to feedback to the \textit{child agent}, the resource-allocating agent, the value of the child agents' allocation strategy to the parents' broader goals. A child agent learns the value of its allocation strategy across many parent agents as part of each parents' overall set of resource demands or tasks. This passes information to the child agent on how its allocation may have effected other child agents' resource allocations more broadly in the system. Through this extension of a localised learning approach we are able to model the complex outcomes involved in multiple resource allocations in an agent system. However, as each agents' view of the system is still constrained, it scales well to large systems.

\newcommand{\varSystem}{s}
\newcommand{\varParentAgent}[2]{\ifx \\#1\\ pg \else pg_{#1}^{#2} \fi}
\newcommand{\setParentAgent}[2]{\ifx \\#1\\ PG \else PG_{#1}^{#2} \fi}
\newcommand{\varChildAgent}[2]{\ifx \\#1\\ cg \else cg_{#1}^{#2} \fi}
\newcommand{\setChildAgent}[2]{\ifx \\#1\\ CG \else CG_{#1}^{#2} \fi}
\newcommand{\powerSetParentAgent}{2^{PG}}
\newcommand{\varAtomicTaskType}[2]{\ifx \\#1\\ tp \else tp_{#1}^{#2} \fi}
\newcommand{\setAtomicTaskType}[2]{\ifx \\#1\\ TP \else TP_{#1}^{#2} \fi}
\newcommand{\varCompositeTaskType}[2]{\ifx \\#1\\ \hat{tp} \else \hat{tp}_{#1}^{#2} \fi}
\newcommand{\setCompositeTaskType}[2]{\ifx \\#1\\ \hat{TP} \else \hat{TP}_{#1}^{#2} \fi}
\newcommand{\powerSetTaskTypes}{2^{\setAtomicTaskType{}{}}}
\newcommand{\varCapabilityMap}{q}
\newcommand{\varAtomicTaskQuality}[2]{\varCapabilityMap_{\varChildAgent{}{}}}

\newcommand{\varTaskGroupMap}{tg}
\newcommand{\varAtomicTask}{t}
\newcommand{\setAtomicTask}{T}
\newcommand{\powerSetTasks}{2^T}
\newcommand{\varAtomicTaskInstanceDetails}{\pi}
\newcommand{\varCompositeTask}{\hat{t}}
\newcommand{\setCompositeTask}{\hat{T}}
\newcommand{\atomicTaskTypeFunction}{type_a}
\newcommand{\compositeTaskTypeFunction}{type_c}
\newcommand{\varFrequencyFunction}{tf}
\newcommand{\varResource}[2]{\ifx \\#1\\ r \else r_{#1}^{#2} \fi}
\newcommand{\setResource}{R}
\newcommand{\varResourceMap}{ar}
\newcommand{\setResourceAssignment}{2^{\setResource{}{} \times \mathbb{R}_{>=0}}}
\newcommand{\varTaskAllocation}{tl}
\newcommand{\setPossibleAllocation}{2^{\setChildAgent{}{} \times \setAtomicTask{}{}}}
\newcommand{\varComponentTasksResult}{ctr}
\newcommand{\powersetTaskResult}{2^{\setAtomicTask{}{} \times \mathbb{R}}}
\newcommand{\varTaskAllocationQuality}{taq}
\newcommand{\varComponentTasksValue}{ctv}
\newcommand{\varResourceWeighting}[2]{\ifx \\#1\\ w \else w_{#1}^{#2} \fi}
\newcommand{\setResourceWeighting}{W}
\newcommand{\varResourceAllocation}{ra}
\newcommand{\setResourceAllocation}{RA}
\newcommand{\sysResourceAllocation}{sra}
\newcommand{\varTime}{\phi}
\newcommand{\setTime}{\Phi}
\newcommand{\absoluteTaskValue}{atv}
\newcommand{\functionAbsoluteTaskValue}[2]{
	\functionSignature{\absoluteTaskValue{}{}}{\varCompositeTask{#1}{}, \varAtomicTask{#2}{}}
}
\newcommand{\formalResourceAllocation}[2]{
	\functionFormal{\varResourceAllocation_{\varChildAgent{}{}}{}}{\setAtomicTask{}{}}{\setResourceAssignment{}{}}
}
\newcommand{\functionResourceAllocation}[2]{
	\functionSignature{\varResourceAllocation_{\varChildAgent{}{}}{}}
	{\varAtomicTask{}{}}
}
\newcommand{\formalResourceMap}[2]{
	\functionFormal{\varResourceMap{}{}}
	{\setChildAgent{}{} \times \setResource{}{}}
	{\setRealNumbersNonNegative{}{}}
}
\newcommand{\functionResourceMap}[2]{
	\functionSignature{\varResourceMap{}{}}
	{\varChildAgent{}{}, \varResource{}{}}	
}

\newcommand{\varSystemUtility}{u}
\section{Resource allocation in a multi-agent system}
\label{section:problem}
In this section we define the problem of resource allocation in dynamic systems and introduce the key notation and definitions to be used subsequently. These are summarised in Appendix \ref{section:symbols}.

Our assumed system is one in which a set of \emph{agents} perform \emph{tasks}. Tasks are typed, and are either \emph{atomic} if they cannot be decomposed into smaller tasks, or \emph{composite} if they comprise a set of atomic tasks. Composite tasks are received by some agents, \emph{parent agents}, throughout the system's lifetime and each composite task is decomposed into its atomic task parts, which are allocated to other agents, \emph{child agents}, to complete. For simplicity, we discuss parent agents and child agents as if distinct groups below, but this is not a requirement, as they are just roles played with regards to composite tasks. Completing tasks requires \emph{resources}, e.g.\ computing power, and each agent has a constant amount of each resource that it allocates to tasks of each given type.

\definition[System]{
	A \textit{system} is a tuple $\varSystem{}{} = (\setParentAgent{}{}, \setChildAgent{}{}, \setAtomicTaskType{}{}, \setCompositeTaskType{}{}, \setResource{}{}, \varResourceMap{}{}, \varTaskGroupMap{}{})$, where
	\begin{itemize}
	    \item $\setParentAgent{}{}$ is a set of parent agents,
	    \item $\setChildAgent{}{}$ is a set of child agents,
	    \item $\setAtomicTaskType{}{}$ is a set of atomic task types,
	    \item $\setCompositeTaskType{}{} \subseteq \powerSetTaskTypes{}{}$ is the set of composite task types (sets of atomic task types) that occur in the system,
	    \item $\setResource{}{}$ is a set of resources needed to perform tasks,
	    \item $\formalResourceMap{}{}$ is a mapping from each child agent and each resource to the amount of that resource that the agent possesses.
	    \item $\varTaskGroupMap{}{} \colon \setCompositeTaskType{}{} \to \powerSetParentAgent{}{}$ is a mapping from each composite task type to the group of parent agents that receive and ensure the completion of tasks of that type.
	\end{itemize}
}

Every atomic task is typed by one of the atomic task types of the system. Each task of the same type is a distinct instance, so completing one task of a type does not complete other tasks of the same type. A composite task is a set of atomic tasks, with no repetition of atomic task types. We denote the set of all possible atomic tasks in a system $\setAtomicTask$, and the set of all possible composite tasks $\setCompositeTask$.

\definition[Atomic task]{
	An \textit{atomic task} in a system is a tuple $\varAtomicTask{}{} = (\varAtomicTaskType{}{}, \varAtomicTaskInstanceDetails{}{})$, where $\varAtomicTaskType{}{} \in \setAtomicTaskType{}{}$ is the type of the task, and $\varAtomicTaskInstanceDetails$ is the specification of this specific task (not defined further here).
}

\definition[Composite task]{
	A \textit{composite task} in a system is a set $\varCompositeTask{}{} \subseteq \setAtomicTask{}{}$.
}

The type of a composite task is the set of types of its elements, and is a member of $\setCompositeTaskType{}{}$. For convenience, we define for atomic tasks, $\atomicTaskTypeFunction{}{} \colon \setAtomicTask{}{} \to \setAtomicTaskType{}{}$ where $\atomicTaskTypeFunction((\varAtomicTask{}{}, \varAtomicTaskInstanceDetails{}{})) = \varAtomicTaskType{}{}$, and for composite tasks, $\compositeTaskTypeFunction \colon \setCompositeTask{}{} \to \powerSetTasks{}{}$ where $\compositeTaskTypeFunction(\{\varAtomicTask_{1}{}, \varAtomicTask_{2}{}, ...,  \varAtomicTask_{n}{}\}) = \{\atomicTaskTypeFunction(\varAtomicTask_{1}{}), \atomicTaskTypeFunction(\varAtomicTask_{2}{}), ..., \atomicTaskTypeFunction(\varAtomicTask_{n}{})\}$.

Tasks of each composite type arrive in the system with a given frequency.

\definition[Task frequency]{
    The frequency at which composite tasks arrive in a system (are received by parent agents to enact) is given by $\varFrequencyFunction{}{} \colon \setCompositeTaskType{}{} \to \mathbb{Z}$, a mapping from each composite task type to the number of time steps between each occurrence of a task of that type being received.
}

\example[Task allocation]{
	A vehicle in a V2X system may receive a composite task of type $\varCompositeTaskType{}{}$ composed of atomic task types $\varAtomicTaskType{}{}_{pos}, \varAtomicTaskType{}{}_{vel}, \varAtomicTaskType{}{}_{con}$ to ask other vehicles for position, velocity, or congestion data for a location. This parent agent may then distribute these atomic tasks to other nearby vehicles, which then act as child agents to it, completing the allocated atomic tasks. These vehicles may be acting as child agents to other vehicles in the system, and have multiple allocated atomic tasks to complete at any given time as a result.
}

The system is capable of processing the tasks in the following way. We assume tasks are independent and have no necessary order of completion.

\begin{enumerate}
	\item A composite task arrives in the system, allocated to the parent agent that ensures completion of tasks of that type.
	\item The parent agent decomposes the composite task into atomic tasks.
	\item The parent agent allocates these atomic tasks to child agents.
	\item A child agent allocated an atomic task performs it using the resources it has available.
	\item Once all the atomic tasks have been completed the composite task is complete.
\end{enumerate}

The quality of result of an atomic task depends on the amount of resources an agent has allocated to performing the task.


\newcommand{\functionAtomicTaskQuality}[2]{
	\functionSignature{\varCapabilityMap_{\varChildAgent{}{}}{}}
	{\varAtomicTaskType{}{},(\varResource{}{}, x){}{} }
}
\newcommand{\functionAtomicTaskQualitySet}[2]{
	\functionSignature{\varCapabilityMap_{\varChildAgent{}{}}{}}
	{\setAtomicTaskType{}{},\setResource{}{} \times \setRealNumbersNonNegative{}{} }
}
\definition[Atomic task quality]{
	\label{def:task_quality}
	The \textit{quality} produced by child agent $\varChildAgent{}{}$ for a task of a given atomic task type it performs using a given amount of each resource is determined by a function $\label{eqn:atomic_task_quality}\varCapabilityMap_{\varChildAgent{}{}} \colon \setAtomicTaskType{}{} \times \setResourceAssignment{}{} \to \mathbb{R}$.
}

Each atomic task in a composite task will give some \emph{value} to the outcome of that composite task, corresponding to how significantly it contributed relative to other atomic tasks. This is not the same as result quality, e.g.\ an atomic task may have been performed to exceptionally high quality but produced results that were replicated by other successful tasks whereas another task performed at lower quality may have produced an important and distinct product/finding. In general, the value of each atomic task to a composite task may be unknown until the composite task is completed.

\newcommand{\functionComponentTasksValue}[2]{
\ifx \\#1\\
	\functionSignature{\varComponentTasksValue{}{}}
	{\varAtomicTask{}{}}
\else
	\functionSignature{\varComponentTasksValue{}{}}
	{#1}
\fi
}
\newcommand{\functionComponentTasksValueSet}[2]{
	\functionSignature{\varComponentTasksValue{}{}}
	{\setAtomicTask{}{}}
}
\definition[Component tasks proportional value]{
	The \textit{proportional value} of each component atomic task of a composite task $\varCompositeTask{}{}$ is expressed by a mapping from the atomic tasks comprising $\varCompositeTask{}{}$ to the fractional value each atomic task contributed to $\varCompositeTask{}{}$ as judged after $\varCompositeTask{}{}$'s completion,    $\varComponentTasksValue{}{} \colon \setAtomicTask{}{} \to \mathbb{R}$, where $\sum\limits_{\varAtomicTask{}{} \in \setAtomicTask{}{}} \varComponentTasksValue(\varAtomicTask{}{}) = 1$ and the value of an atomic task to a composite task of which it was not part is zero.
}

The aim of individual agents in the system is to ensure those atomic tasks that are of most value to a composite task are performed to the highest quality, thus ensuring the composite tasks produce the best results. For child agents enacting atomic tasks, this means allocating more resources to high value tasks.

There are obstacles to this goal. First, the value of the atomic task to the composite task may not be known in advance, as said above. Second, a given child agent cannot know when it will receive atomic tasks to enact because, even if the frequency by which composite tasks are received by the system is constant and known, it is a choice of a parent agent as to who to allocate atomic tasks to in any given instance. If a child agent has already allocated resources to one atomic task under execution, then it cannot use those resources for a new task of potentially higher value.

To address this latter point, we assume that each child agent allocates in advance a portion of its resources to each atomic task type. We are agnostic as to whether tasks are then executed sequentially or in parallel, which will vary per application. An agent's resource allocation can change over time: specifically, the solutions presented later in this paper allow an agent to learn a good resource allocation for the tasks it is assigned and the value they have to composite tasks.

\definition[Agent resource weighting]{
	The \textit{resource weighting} of a child agent $\varChildAgent{}{}$ at time $\varTime{}{}$ is a mapping from each atomic task type and each resource to the proportion of that resource it possesses which it devotes to tasks of that type, $\varResourceWeighting{}{}_{\varChildAgent{}{}, \varTime{}{}} \colon \setAtomicTaskType{}{} \times \setResource{}{} \to \mathbb{R}$, where $\forall \varChildAgent{}{} \in \setChildAgent{}{}, \varTime \in \mathbb{Z}, \varResource{}{} \in \setResource{}{}, \sum\limits_{\varAtomicTaskType{}{} \in \setAtomicTaskType{}{}} \varResourceWeighting{}{}_{\varChildAgent{}{}, \varTime{}{}} (\varAtomicTaskType{}{}, \varResource{}{}) = 1$.
}

\definition[Agent resource allocation]{
    The \textit{resource allocation} of a child agent $\varChildAgent{}{}$ at time $\varTime{}{}$ is a mapping from each atomic task type to the amount of each resource it devotes to that type at that time, $\varResourceAllocation_{\varChildAgent{}{}} \colon \setAtomicTask{}{} \to \setResourceAssignment{}{}$ such that $\varResourceAllocation_{\varChildAgent{}{}}(\varAtomicTask{}{}) = \{ (\varResource{}{}, \varResourceWeighting{}{}_{\varChildAgent{}{}, \varTime{}{}}(\varAtomicTask{}{}, \varResource{}{})\varResourceMap(\varChildAgent{}{}, \varResource{}{})) | \varResource{}{} \in \setResource{}{} \}$.
}

\definition[System resource allocation]{
    The set of all agents' resource allocations in the system at time $\varTime{}{}$ is specified by $\sysResourceAllocation_{\varTime{}{}} \colon \setChildAgent{}{} \times \setAtomicTask{}{} \to \setResourceAssignment{}{}$.
}

\example[Resource weighting]{
	A child agent $\varChildAgent{}{}$ in a vehicle-to-everything (V2X) communication system is processing atomic tasks to provide its position $\varAtomicTaskType{}{}_{pos}$ and congestion information $\varAtomicTaskType{}{}_{con}$ to parent agents $\setParentAgent{}{}$. Both task-types require the child agents' memory resource $\varResource{mem}{}$. If the parent agents are close to the child agent, then positional updates are more important than congestion information to them, and the child agents' resource weightings $\varResourceWeighting{mem}{}(pos) >> \varResourceWeighting{mem}{}(con)$. However, if the parent agents are far away, congestion information is more valuable to them than position updates for $\varChildAgent{}{}$, and so resource weights should be skewed towards the congestion task $\varResourceWeighting{mem}{}(pos) << \varResourceWeighting{mem}{}(con)$.
}

Which child agents will carry out the atomic tasks of a composite task are determined by the task allocation of the parent agent enacting the composite task.

\definition[Task allocation]{
	A \textit{task allocation} is a mapping from atomic tasks to the child agents that are allocated to perform them, $\varTaskAllocation{}{} \colon \setAtomicTask{}{} \to \setChildAgent{}{}$.
}

The set of results of atomic tasks are returned to the parent agent enacting their encapsulating composite task. From the task allocation, resource allocation, and quality function, we can determine the resulting qualities of these tasks. Given the above definitions, we can say that the quality of the result of an atomic task $\varAtomicTask{}{}$ performed at time $\varTime{}{}$ through a given task allocation $\varTaskAllocation{}{}$ and a given system resource allocation at that time $\sysResourceAllocation_{\varTime{}{}}$ is determined by $\varCapabilityMap_{\varChildAgent{}{}} (\psi, \sysResourceAllocation_{\varTime{}{}} (\varChildAgent{}{}, \psi))$ where $\varChildAgent{}{} = \varTaskAllocation(\varAtomicTask{}{})$ and $\psi = \atomicTaskTypeFunction(\varAtomicTask{}{})$.

The quality of result of a composite task depends on the qualities of its component atomic tasks and the value of each to the composite task, i.e.\ it is higher where the most important tasks were performed well.

\definition[Composite task quality]{
	The \textit{quality} of composite task $\varCompositeTask{}{}$ whose atomic components were performed at time $\varTime{}{}$ under task allocation $\varTaskAllocation{}{}$ and system resource allocation $\sysResourceAllocation_{\varTime{}{}}$ is $\varTaskAllocationQuality_{\varTime{}{}}(\varCompositeTask{}{}) = \sum\limits_{\varAtomicTask{}{} \in \varCompositeTask{}{}} \varComponentTasksValue(\varAtomicTask{}{})\varCapabilityMap_{\varChildAgent{}{}} (\psi, \sysResourceAllocation_{\varTime{}{}} (\varChildAgent{}{}, \psi))$ where $\varChildAgent{}{} = \varTaskAllocation(\varAtomicTask{}{})$ and $\psi = \atomicTaskTypeFunction(\varAtomicTask{}{})$.
}

From this we can define, first, the absolute value of each atomic task to the system being the product of the composite task's quality and the atomic tasks relative value in generating that quality and, second, the utility of the system during a time period as being the total qualities of the composite tasks it executes in that period.

\definition[Component tasks absolute value]{
    The \textit{absolute value} of each component atomic task of a composite task $\varCompositeTask$ executed at time $\varTime$ is expressed by a mapping from the atomic tasks comprising $\varCompositeTask$ to the actual value provided by that task according to the quality of the composite task and the proportional value of the atomic task within the composite, $\absoluteTaskValue \colon \setCompositeTask \times \setAtomicTask \to \mathbb{R}$, such that $\absoluteTaskValue(\varCompositeTask, \varAtomicTask) = \varTaskAllocationQuality_{\varTime}(\varCompositeTask)\varComponentTasksValue(\varAtomicTask)$
}

\definition[System utility]{
    The utility of a system in the period $\setTime$ is given by $\varSystemUtility(\setTime) = \sum\limits_{\varTime \in \setTime} \sum\limits_{\varCompositeTask \in \setCompositeTask_{\varTime}} \varTaskAllocationQuality_{\varTime}(\varCompositeTask)$, where $\setCompositeTask_{\varTime}$ is the set of composite tasks enacted at time $\varTime$.
}

\example[Task value]{
Vehicles $\varChildAgent{1}{}$ and $\varChildAgent{1}{}$ complete and return data for the allocated congestion tasks $\varAtomicTask_{1}$ and $\varAtomicTask_{2}$ respectively for the same parent agent $\varParentAgent{}{}$ as part of its composite task $\varCompositeTask{}{}$. The atomic task quality for a vehicles' congestion task type is greater the more recent the data is. However, if $\varAtomicTask_{2}$ contains more recent data than $\varAtomicTask_{1}$, or duplicates its information, then $\varAtomicTask_{1}$ has less value to $\varCompositeTask{}{}$ than if only $\varAtomicTask_{1}$ were part of the composite task, and so $\functionComponentTasksValue{\varAtomicTask{}{}_{1}}{}$ returns a lower component task absolute value. 
}
\newcommand{\formalPAG}{pag_{\varChildAgent{}{}}}
\newcommand{\functionPAG}[2]{
	\ifx \\#1\\
		\formalPAG(\varParentAgent{}{})
	\else
		\formalPAG(#1)
	\fi
}
\newcommand{\functionPAGSet}[2]{
\ifx \\#1\\
	\formalPAG(\setParentAgent{}{})
\else
	\formalPAG(\setParentAgent{#1}{#2})
\fi
}

\newcommand{\formalPGroup}[2]{
	\functionFormal{pgroup_{\varChildAgent{}{}}}{\setParentAgent{}{}}{\powerSetParentAgent{}{}}
}
\newcommand{\functionPGroup}[2]{
	\ifx \\#1\\
	\functionSignature{pgroup_{\varChildAgent{}{}}}{\varParentAgent{}{}}
	\else
	\functionSignature{pgroup_{\varChildAgent{}{}}}{#1}
	\fi
}
\newcommand{\functionPGroupSet}[2]{
	\functionPGroup{\setParentAgent{}{}}{}
}
\newcommand{\formalPAGResourceWeighting}{
	\functionFormal
	{pgw_{\varChildAgent{}{}, \varResource{}{}, \varTime{}{}}}
	{\functionPGroupSet{}{} \times \setAtomicTaskType{}{}}
	{ \mathbb{R}}
}

\newcommand{\functionPAGResourceWeighting}[2]{
	\functionSignature
	{pgw_{\varChildAgent{}{}, \varResource{}{}, \varTime{}{}}}
	{\setParentAgent{#1}{}, \setAtomicTaskType{#2}{}}
}
\newcommand{\setResourceWeight}[2]{
	\setSymbol{W}{\varChildAgent{}{}, \varResource{}{}}{#2}
}
\newcommand{\matrixResourceWeight}[2]{
	\matrixSymbol{PW}{\varChildAgent{}{}, \varResource{}{}}{#2}
}
\newcommand{\varPAGResourceWeight}[2]{
	{pw}_{\setParentAgent{#1}{}, \varAtomicTaskType{#2}{}}
}
\newcommand{\setPAGResourceWeight}[2]{
	{pw}_{\setParentAgent{#1}{}, \setAtomicTaskType{#2}{}}
}
\newcommand{\varParentTaskIndex}{pti_{\varChildAgent{}{}}}
\newcommand{\formalFunctionParentTaskIndex}{
	\functionFormal{\varParentTaskIndex{}{}}
	{\setParentAgent{}{} \times \setAtomicTaskType{}{}}
	{\setIntegersPositive{}{} \times \setIntegersPositive{}{}}
}
\newcommand{\functionParentTaskIndex}{
	\functionSignature
	{\varParentTaskIndex{}{}}
	{\varParentAgent{}{}, \varAtomicTaskType{}{}}
}

\newcommand{\formalPAGSampleCount}{
	\functionFormal{pags_{\varChildAgent{}{}, \varTime{}{}}{}}
	{\setParentAgent{}{}}
	{\setIntegersNonNegative{}{}}
}
\newcommand{\functionPAGSampleCount}[2]{
	\ifx \\#1\\
	\functionSignature{pags_{\varChildAgent{}{},\varTime{}{}}{}}
	{\setParentAgent{}{}}
	\else
	\functionSignature{pags_{\varChildAgent{}{},\varTime{}{}}{}}
	{#1}
	\fi
}
\newcommand{\varKullbackLiebler}{dkl}
\newcommand{\functionKullbackLiebler}[2]{\varKullbackLiebler(\setParentAgent{#1}{})}
\newcommand{\setWeightBlending}{B_{\varChildAgent{}{}}}

\providecommand{\functionSumNormSymbol}[2]{\texttt{norm}{#1}{#2}}
\providecommand{\functionSumNormSignature}[2]{
	\ifx \\#1\\
	\functionSignature{\functionSumNormSymbol{}{}}
	{\setX{}{}}
	\else
	\functionSignature{\functionSumNormSymbol{}{}}
	{#1}
}
\newcommand{\functionSumNormRowSymbol}[2]{\texttt{norm}{#1}{#2}}
\newcommand{\functionSoftmaxSymbol}[2]{\varSymbol{\sigma}{#1}{#2}}
\newcommand{\functionSoftmaxSignature}[2]{
	\ifx \\#1\\
	\functionSignature{\functionSoftmaxSymbol{}{}}
	{\varX{}{}}
	\else
	\functionSignature{\functionSoftmaxSymbol{}{}}
	{#1}
}
\newcommand{\functionSoftmaxSet}[2]{
	\sigma(X) = \lbrace \frac{e^{x_i}}{\sum_{j=1}^{\funcSize{X}} e^{x_j}}\rbrace_{\forall x_i \in X}
}

\newcommand{\varCombinedResourceWeights}{crw}
\newcommand{\formalFunctionCombinedResourceWeights}{crw_{\varChildAgent{}{}}}
\newcommand{\formalFunctionCombinedResourceWeightsSignature}{\formalFunctionCombinedResourceWeights(\setWeightBlending{}{}, \matrixResourceWeight{}{})}
\newcommand{\functionCombinedResourceWeightsSignature}{\formalFunctionCombinedResourceWeights(\setWeightBlending{}{}, \matrixResourceWeight{}{})}
\newcommand{\matrixCombinedResourceWeights}{\matrixSymbol{C}}
\newcommand{\varEligibilityTrace}[2]{\varSymbol{e}{#1}{#2}}
\newcommand{\matrixEligibilityTrace}[2]{\matrixSymbol{E}{\varChildAgent{}{}}{#2}}
\newcommand{\functionEligibilityTraceUpdateSignature}[2]{
	\functionSignature{etu}
	{
		\matrixEligibilityTrace{}{}, \varParentAgent{}{},
		\varAtomicTaskType{}{},
		\gamma
	}
}
\section{Learning High Value Resource Allocations}
\label{section:solution}
As defined above, the aim here is to improve the system utility by child agents allocating more resources to tasks that will be of more value within their composite tasks. As explained, a child agent allocates resources to atomic task types in advance of being allocated tasks of those types to complete due to the unpredictable nature of the allocation from its perspective. Our solution is influenced by three problems arising from the problem domain.
\paragraph{Unpredictable task value} The first problem is that the value an atomic task will have may be unknown in advance. However, here we assume that while unpredictable, there may be biases or patterns to be learnt. We further assume that these patterns will correspond to the parent agents from whom atomic tasks are allocated, i.e.\ a given parent agent will typically be performing composite tasks for a given purpose within the system, and so the relative value of each atomic task may have similarities across the composite tasks allocated by that parent agent. Much of the solution below would apply equally should a particular system have another way to categorise incoming tasks, other than the parent agent, relevant to learning their likely value.
\paragraph{Varying allocation frequencies} The second problem is that a parent agent will be making allocation decisions that change which atomic tasks a given child agent receives, and the parent agent's strategy could vary over time. Additionally, the frequencies at which composite tasks of each type arrive in the system may change. This means that a child agent will not consistently receive atomic tasks of different types at the same rate and their arrival will not be uniform. As a child agent can only have one resource allocation, it will not be beneficial to allocate many resources to a task type for which tasks of that type are likely to be high value if arriving from a given parent agent, but that parent agent is not allocating the child any tasks at this time.
\paragraph{Learning resource cost} The final problem is that learning requires resources and child agents have limited resources, as specified in the previous section. Any given system may be large in the number of parent agents. If a child agent attempted to learn patterns in which atomic tasks types are most valuable for every parent agent, this may be infeasible in terms of memory or computational costs.

The solution we propose uses reinforcement learning to adapt resource weightings for child agents given the composite task values of the atomic tasks it receives from parent agents. We take the approach summarised below.
\begin{enumerate}
\item Each child agent will allocate an amount of resources to learning parent agent task value patterns. In this paper, the amount will be fixed in advance, and we leave varying the balance of resources allocated to learning versus task execution to future work. When referring to the resources possessed by an agent, we will exclude the resources allocated to learning for simplicity.
\item A child agent groups parent agents in the system and treats each group as if it was one parent agent for the purposes of learning. The number of agents per group is dictated by the resources allocated to learning: more resources allows for more, smaller groups.
\item For each parent agent group, there is a learnt model of the value of each atomic task type. When a parent agent completes a composite task, the child agents that completed the corresponding atomic tasks are sent the absolute values for those tasks. The child agent then adjusts its model for the relevant parent-agent group. 
\item The child agent regularly aggregates the learnt models to determine a resource weighting across atomic task types and adjusts its resource allocation accordingly for subsequent tasks.
\end{enumerate}

\subsection{Grouping parent agents to constrain resources}
As learning requires the use of limited resources, a solution that models the value of resource allocation for each parent agent individually would be limited in scalability, as the resources required for learning increase with the number of parent agents. In contrast, if the solution used the incoming atomic tasks for all parent agents and modelled them as-one, it would be unlikely to provide optimal results where there were many agents making requests, due to the problems described previously. For example, if a parent agent $\varParentAgent{1}{}$ made a request for resources at time $\varTime{1}{}$, then a number of other parent agents made requests, followed by $\varParentAgent{1}{}$ again at time $\varTime{2}{}$, there would be many changes to the model in-between each request by $\varParentAgent{1}{}$. Therefore the child agents' resource allocation at $\varTime{2}{}$ is unlikely to be close to optimal for the demands of $\varParentAgent{1}{}$, the information on optimal values for $\varParentAgent{1}{}$ is effectively lost between $\varTime{1}{}$ and $\varTime{2}{}$.

To solve this problem we combine parent agents into groups. Tasks allocated from each group of parent agents are treated as the same distribution for resource allocation by the child agent. Resource weightings are similarly defined for each parent agent group rather than for each parent agent individually. This approach allows us to constrain the usage of resources for modelling by the child agent, while retaining information on parent agents requests over longer time periods. To simplify notation we combine each of the parent group weightings into a matrix.
\definition[Parent agent group]{
	A \textit{parent agent group} is a fixed mapping for each child agent to sets of parent agents that are grouped together for task value modelling,
	$\functionFormal{\formalPAG{}{}}{\setParentAgent{}{}}{\powerSetParentAgent{}{}}$. For convenience, we define the set of $m$ groupings of parent agents $\setParentAgent{}{}$ as $\formalPGroup{}{}$ where $\functionPGroup{\setParentAgent{}{}}{} =
	\lbrace
		\setParentAgent{1}{}, \dots, \setParentAgent{m}{}
	\rbrace$ such that $
	\functionPAG{\setParentAgent{i}{}}{} = \setParentAgent{i}{}
	$.
}

\definition[Parent group weights]{
	The \textit{parent group weights} define a set of resource weights for each parent agent group for an individual resource $\varResource{}{}$,
	$\formalPAGResourceWeighting{}{}$. For brevity, for $\setParentAgent{i}{} \in \functionPGroup{\setParentAgent{}{}}{}$, and an atomic task type $\varAtomicTaskType{j}{}$, we use the shorthand notation
	$
	\varPAGResourceWeight{i}{j}
	= \functionSignature
		{pgw_{\varChildAgent{}{}, \varResource{}{}, \varTime{}{}}}
		{
			\functionPAG{\setParentAgent{i}{}}{}, \varAtomicTaskType{j}{}
	}
	$
}

\definition[Parent group weights matrix]{
	For $m$ parent agent groups $\lbrace \setParentAgent{1}{}, ... ,\setParentAgent{m}{}\rbrace$, with $n$ atomic task types $\lbrace \varAtomicTaskType{1}{}, ... ,\varAtomicTaskType{n}{} \rbrace$, we define the \textit{parent group weights matrix} as,
\begin{equation}
\matrixResourceWeight{}{} =
\begin{bmatrix}
	\varPAGResourceWeight{1}{1} & \dots  & \varPAGResourceWeight{1}{n} \\
	\varPAGResourceWeight{2}{1} & \dots  & \varPAGResourceWeight{2}{n} \\
	\vdots & \dots   & \vdots \\
	\varPAGResourceWeight{m}{1} & \dots & \varPAGResourceWeight{m}{n}
\end{bmatrix}
\end{equation}
}

\subsection{Combining parent agent group resource allocation models}

Although a child agent models resource allocations for each of its parent agent group individually, we assume its real resource allocation cannot be instantly changed, and so the child agent must choose a single model, an aggregate of all of its parent agent groups' models, to apply for each resource it owns. 
\example[Resource re-allocation]{
	A car in a V2X system may allocate some battery power to its wireless transmitter. However, to reallocate this resource to another component such as a LIDAR (Light Detection and Ranging) system, would involve a delay in effecting the component, and is not instantaneous.
}
To do this we define a function to combine each of the separate parent group models and give a single model that is used for actual resource allocation. This function uses the frequency of atomic task types for each parent agent group, and the entropy of their corresponding resource weights, to best combine the models. The reasons for this method of combination are described below. 

\subsubsection*{Task allocation frequency and resource allocation impact.}
In a situation where we have two sets of parent agents groups $\setParentAgent{1}{}, \setParentAgent{2}{} \in \functionPGroupSet{}{}$, we will have corresponding distributions of resource weights $\setPAGResourceWeight{1}{}$ and $\setPAGResourceWeight{2}{}$ respectively. If the frequency of incoming tasks from group $\setParentAgent{1}{}$ is significantly greater than that of $\setParentAgent{2}{}$, we want the child agents' actual resource allocation to be closer to the values in $\setPAGResourceWeight{1}{}$ as we will generate more system utility through applying that allocation given the overall distribution of incoming tasks. To let the preferred resource allocation of some parent agent groups contribute to the child agents final resource allocation more strongly than that of others we utilise sample counting to measure the relative frequency of all tasks received from each parent agent group.
\definition[Parent agent sample count]{
	The \textit{parent agent sample count} maps groups of parent agents $\setParentAgent{}{}$ of a child agent $\varChildAgent{}{}$ to a count of the number of times any task has been allocated to the child agent from a member of that group up to time $\phi$, $\formalPAGSampleCount{}{}$
}

\subsubsection*{Using the relative entropy of resource weightings to measure resource allocation policy impacts.}
Low entropy in resource weightings across a parent group can mean that either,
\begin{itemize}
	\item {
		the overall outcome of the combined tasks in that group are not strongly influenced by the resource allocation policy of the child agent
	}
	\item {
		allocating resources to the atomic tasks of that group has a similar impact on the quality of the composite tasks of each of the parent agents in that group.
	}
	\item {
		the distribution of incoming tasks for that is more dynamic or unpredictable, possible due to the parent agents changing allocations or them receiving varying composite tasks
	}
\end{itemize}
We therefore use the principle of maximum entropy of resource allocation \cite{Johansson2005} to increase the influence of the resource weightings of parent agent groups that,
\begin{itemize}
 	\item {
 		are strongly impacted by the child agents' resource allocation policy.
 	}
 	\item {
 		where the atomic tasks the parent agents are allocating to the child agent do not give atomic task qualities that are very similar.
 	}
 	\item {
 		provide stable distributions of atomic task allocations to the child agent \cite{Rygielski2011,Tomczak2016}. 
	}
\end{itemize}
\definition[Parent group weights entropy]{
	Given a parent agent group $\setParentAgent{}{}$, with $n$ atomic task types $\lbrace \varAtomicTaskType{1}{}, \dots, \varAtomicTaskType{n}{} \rbrace$ then the \textit{parent group weights entropy} is the relative entropy
	\footnote{The relative entropy or Kullback-Liebler divergence of distribution $P$ from the uniform distribution, $U$ is defined as $
		\sum P(x)
		\log \frac {P(x)} {U(x)}$
	} of the parent groups resource weights,
	\begin{equation}
		\functionKullbackLiebler{}{}
		\funcdef
		\sum \limits_{i=1}^{n}
		\varPAGResourceWeight{}{i}
		\log \left(\frac {\varPAGResourceWeight{}{i}} {\setUniform{}{}(i)} \right)
	\end{equation}
}

\subsubsection*{Defining a blending function for aggregating models.}

Child agents aggregate the learnt models of resource allocation for each parent agent group into one model. To do this we define the \textit{resource weights blending function}, a function that balances the relative sample counts and entropy of each parent group as defined previously, then use these to scale and sum the resource weights of each group into a final resource allocation. We use \textit{sum-normalisation} \footnote{Sum-normalisation is defined by the equation $\functionSumNormSymbol{}{}(X) = \lbrace \frac{x_i}{\sum X} \rbrace_{\forall x_i \in X}$} to balance the impacts of entropy and relative sample frequency on the final resource weights.

\definition[Resource weights blending vector]{
	Given a child agent with parent agent groups $\lbrace\setParentAgent{1}{}, \dots, \setParentAgent{m}{} \rbrace$ the \textit{resource weights blending vector} is sum of the sum-normalised vector of its parent-agent group sample counts and the sum-normalised vector of their resource weight entropies.
	\begin{align}
	\label{eqn:resource_weights_blending}
	\setWeightBlending{}{}
	=& \ 
	\functionSumNormRowSymbol{}{}(
		\begin{bmatrix}
			\functionKullbackLiebler{1}{} & \dots & \functionKullbackLiebler{m}{} 
		\end{bmatrix}^T
	) \nonumber \\
	 & +  
	\functionSumNormRowSymbol{}{}(
	\begin{bmatrix}
			\functionPAGSampleCount{\setParentAgent{1}{}}{} & \dots & \functionPAGSampleCount{\setParentAgent{m}{}}{} 
	\end{bmatrix}^T
	)
	\end{align}
}

\subsubsection*{Combining resource weights into a single model.}
With the grouping of parent agents' respective resource weights, and a blending matrix based on relative parent-group task frequency and per-group entropy of these resource weights, we can now define a function that will combine the two to give a single resource allocation model for each child agent. 
\definition[Combined resource weights function]{
	The \textit{combined resource weights function} takes the parent-agent groups resource weights matrix $\matrixResourceWeight{}{}$ and the blending matrix $\setWeightBlending{}{}$ as parameters and outputs a vector of resource weights for atomic task types $\lbrace \varAtomicTaskType{1}{}, \dots, \varAtomicTaskType{n}{} \rbrace$. These values are used by the child agent as its resource allocation model to apply to real resources. \textit{Softmax normalisation}\footnote{Softmax normalisation is defined as $\functionSoftmaxSet{}{}$} is used to ensure the final resource weights sum to $1$ as required.
	\begin{equation}
	\label{eqn:combined_resource_weights}
\functionCombinedResourceWeightsSignature
\funcdef \functionSoftmaxSignature{
	\functionSumNormSignature{
		\setWeightBlending^T}{} \matrixResourceWeight{}{}
}{}
	\end{equation}
}

\example[Model aggregation]{
	A vehicle $\varChildAgent{}{}$ is receiving tasks from other vehicles combined into three separate parent agent groups. Vehicles in $\functionPAG{\setParentAgent{1}{}}{}$ frequently request position and speed updates as they are near to $\varChildAgent{}{}$, giving a high sample count. Those in $\functionPAG{\setParentAgent{2}{}}{}$ are not nearby, but as they are about to enter a congested area, congestion task completions are more valuable to them than others, so their resource weights have a high entropy. Finally, vehicles in $\functionPAG{\setParentAgent{3}{}}{}$ are neither nearby nor nearing congestion so tasks are of relatively equal value to them. In these circumstances, where $\functionPAGSampleCount{\setParentAgent{1}{}}{} >> \functionPAGSampleCount{\setParentAgent{2}{}}{}, \functionPAGSampleCount{\setParentAgent{3}{}}{}$, the combined resource weight function will output an aggregated model that more closely resembles the individual model for $\functionPAG{\setParentAgent{1}{}}{}$, maximising the absolute values returned from highly frequent tasks.
	Similarly, as $\functionKullbackLiebler{\setParentAgent{2}{}}{} >> \functionKullbackLiebler{\setParentAgent{1}{}}{}, \functionKullbackLiebler{\setParentAgent{3}{}}{}$, the combined resource weight function will be closer to that of $\functionPAG{\setParentAgent{2}{}}{}$, to maximise the value returned due to tasks that are more important to their respective parent agents. As the resource model from $\functionPAG{\setParentAgent{3}{}}{}$ is neither frequent, nor generates significantly more value if a particular resource model is applied, it has low sample count and low entropy and so has its impact on the final model reduced in both cases.
}

\subsection{Past resource allocation effects on current task qualities.}
\begin{figure}[t]
\includegraphics[width=\linewidth]{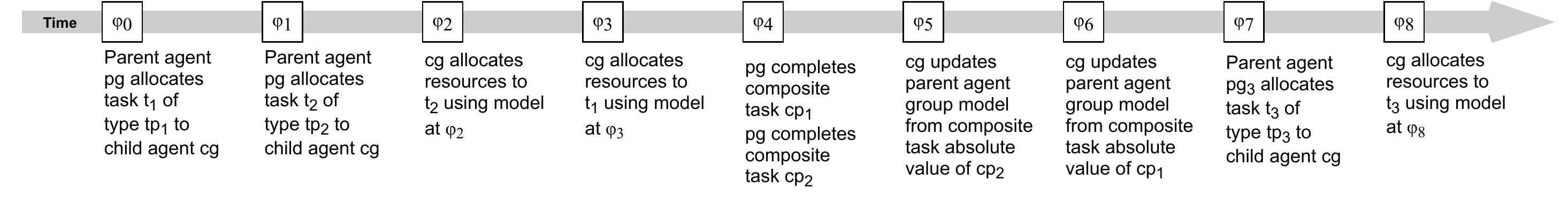}
\caption{Model updates with asynchronous task allocation}
\label{fig:mgrao-eligibility-tracing}
\Description[Parent agents allocate tasks and update models]{The horizontal arrow shows increasing time as parent agent allocate tasks and update their models}
\end{figure}

Over a period of time a child agents' resource allocation model will change as it adapts to the composite task absolute values returned by multiple parent agents. As resources are fixed, each change in resource allocation value for one task type will necessarily change the values for the others. For example, see Figure \ref{fig:mgrao-eligibility-tracing}, where a parent agent $\varParentAgent{}{}$ allocates atomic tasks of type $\varCompositeTaskType{1}{}$ and $\varCompositeTaskType{2}{}$ to a child agent $\varChildAgent{}{}$. On completing the corresponding composite tasks, the parent agent returns the component task absolute values $\functionAbsoluteTaskValue{1}{1}$ and $\functionAbsoluteTaskValue{2}{2}$ to the child agent, which updates its resource allocation model. Next, a new task of type $\varCompositeTaskType{3}{}$ is allocated and completed. When $\functionAbsoluteTaskValue{3}{3}$ is returned there is a problem in deciding which, and how much, each previous model update effected the value of $\varCompositeTaskType{3}{}$. 

This challenge in attributing outcomes to past actions when multiple actions may have contributed to it is an example of the \textit{credit assignment problem} \cite{Sutton1984} commonly found in reinforcement learning systems. To take account of this effect we use a standard technique, a \textit{replacement eligibility trace matrix} \cite{Singh1996}, to apply the effect of absolute task values backwards through past model changes. This enables the \acronymResourceAllocationAlgorithm{}{} algorithm to attribute some of the current model quality to past resource allocation changes.

To allow us to easily map task types and parent-groups into indices in the resource weights matrix or the corresponding eligibility trace matrix we first define an index mapping as follows.

\definition[parent-task type index mapping]{
	Each element in a matrix of $m$ parent agent groups, $\functionPGroupSet{}{} = \lbrace \setParentAgent{1}{}, \dots, \setParentAgent{m}{} \rbrace$, and $n$ atomic task types, $\setAtomicTaskType{}{} = \lbrace \varAtomicTaskType{1}{}, \dots, \varAtomicTaskType{m}{} \rbrace$, is indexed by the \textit{parent-task index} mapping, $\formalFunctionParentTaskIndex{}{}$.
}
\definition[eligibility trace matrix]{
	A child agents' \textit{eligibility trace matrix} $\matrixEligibilityTrace{}{}$ has the same shape as the parent group weights matrix $\matrixResourceWeight{}{}$ with each element initialised to $0$. Given an atomic task type $\varAtomicTaskType{}{}$, a parent agent $\varParentAgent{}{}$, and a fixed decay factor $\gamma$, updates are carried out to each element (i, j) of the eligibility trace matrix using the \textit{eligibility trace matrix update} defined below. 
	\begin{equation}
		\label{eqn:eligibility_trace_update}
		\functionEligibilityTraceUpdateSignature{}{} 
		\funcdef 
		\varEligibilityTrace{ij}{} \leftarrow
		\begin{cases}
			1 & \text{if } (i,j)=\functionParentTaskIndex{}{} \\
			
			\gamma{}{}\varEligibilityTrace{ij}{} & \text{if } \varEligibilityTrace{ij}{} > 0 \\
			
			0 & \text{otherwise} \\
		\end{cases}
	\end{equation}
	The $(i, j)$ element in the eligibility trace matrix corresponding to the $(\functionPAG{\setParentAgent{i}{}}{}, \varAtomicTaskType{j}{})$ element in the parent group weights matrix $\matrixResourceWeight{}{}$ is set to one. All other elements are multiplied by the decay factor $\gamma$. The effect of this is that the eligibility trace matrix measures how recently each task type has been allocated by each parent agent group.
}
\subsection{Using the \textit{\acronymResourceAllocationAlgorithmExtended{}{}} algorithm to maximise system utility}
\label{section:algorithms}
\begin{figure}[ht]
	\centering
	\includegraphics[width=0.7\linewidth]{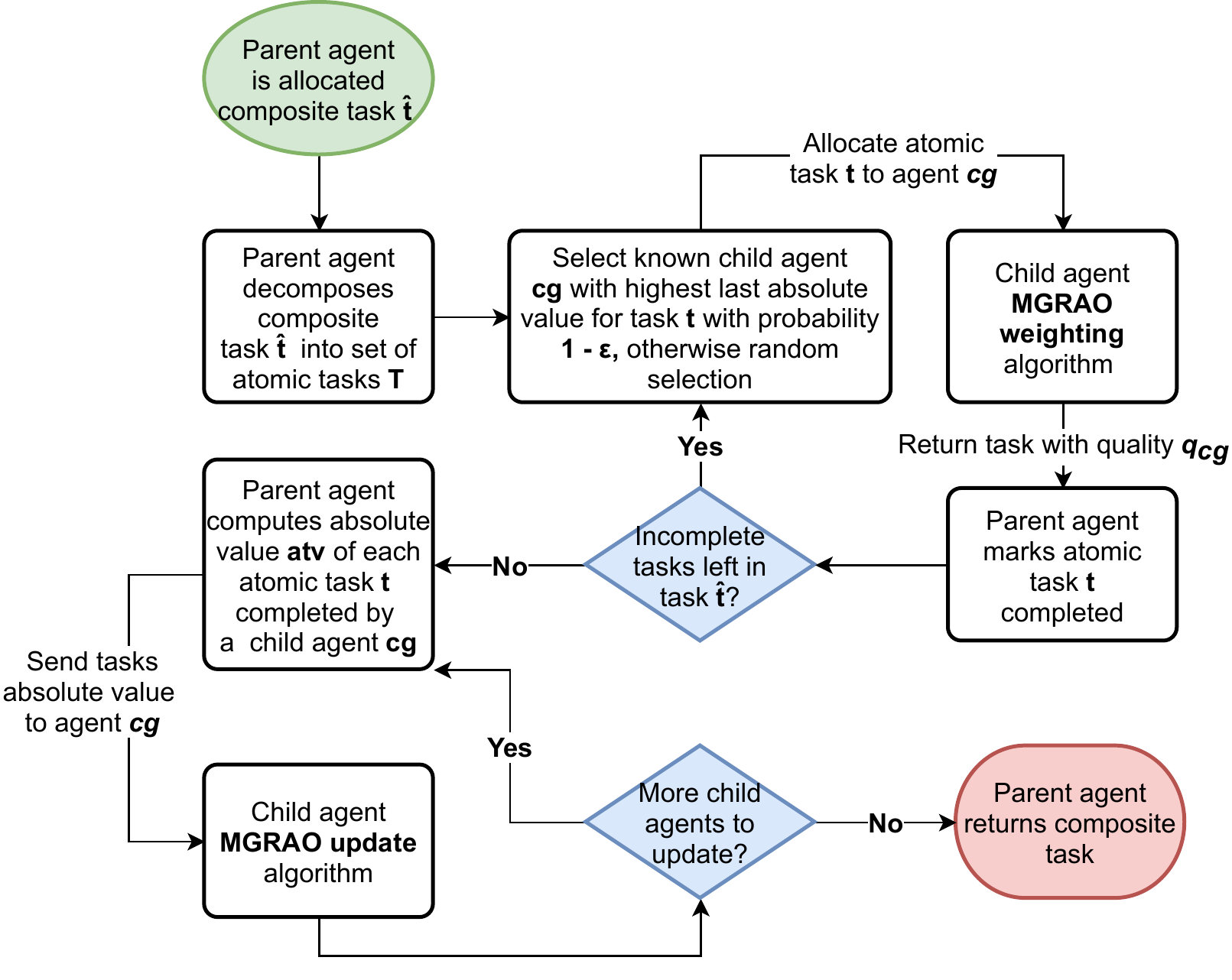}
	\captionsetup{labelfont=bf,singlelinecheck=on}
	\caption{High-level view of the \acronymResourceAllocationAlgorithm{}{} workflow}
	\label{fig:mgrao-highlevel}
	\Description[The flow of the \acronymResourceAllocationAlgorithm{}{} algorithm]{The \acronymResourceAllocationAlgorithm{}{} algorithm flow as it allocates tasks and updates models}
\end{figure}

\begin{figure}[ht]
	\centering
	\includegraphics[width=0.6\linewidth]{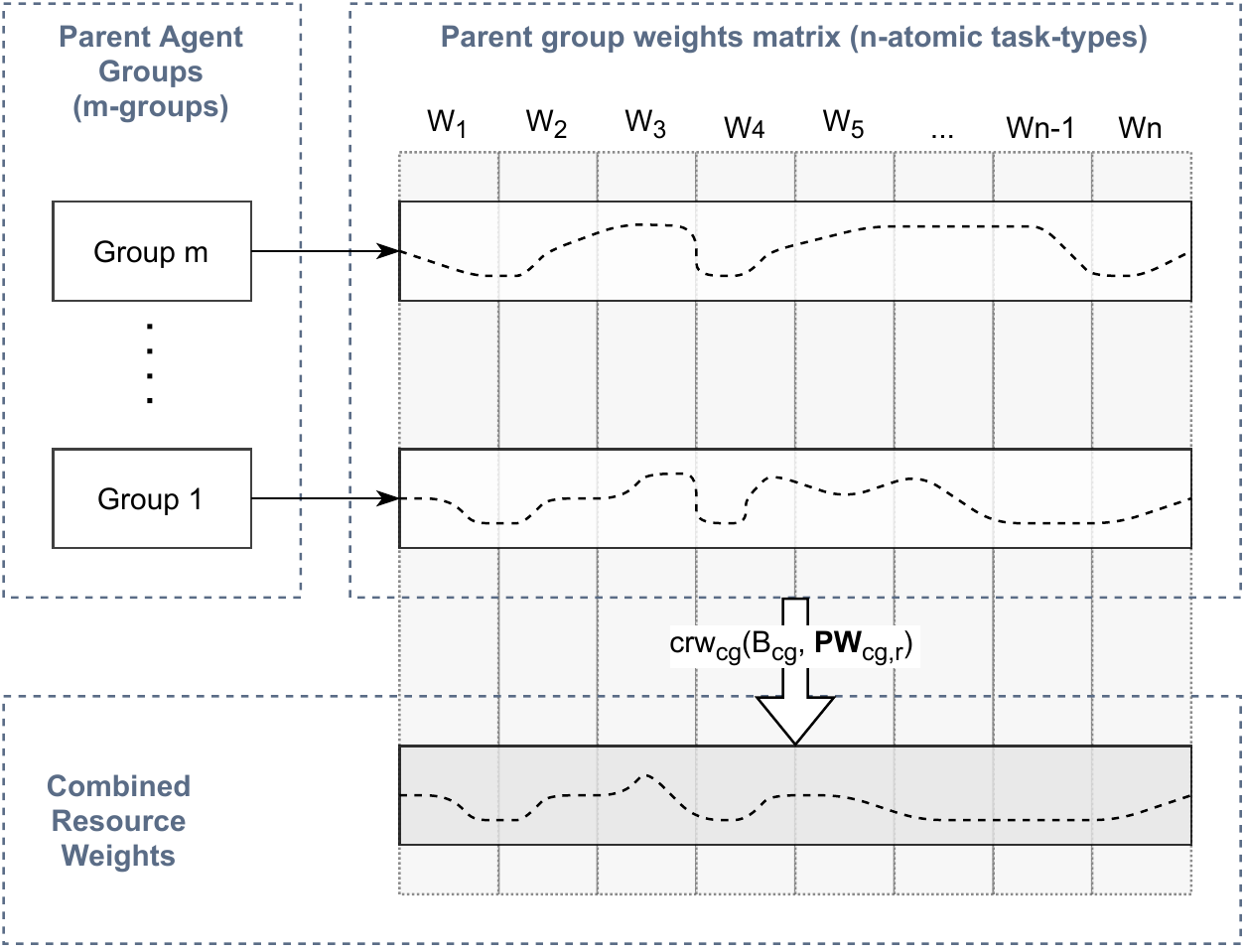}
	\captionsetup{labelfont=bf,singlelinecheck=on}
	\caption{MG-ROA algorithm weight blending}
	\label{fig:mgrao-overview-compressed}
	\Description[The combination of resource models]{Parent agent groups' individual models are combines to give one model to apply to real resources}
\end{figure}

\newcommand{\varReward}[2]{\varSymbol{r}{#1}{#2}}
\newcommand{\varLearningRate}[2]{\alpha}
\newcommand{\tupleResourceAllocationPair}[2]{(\varAgent{#1}{}, \varAtomicTaskType{#2}{})}
With parent groups, resource weights, blending vector, and the combined resource weights function defined, we can bring these together to form the \textit{\acronymResourceAllocationAlgorithmExtended{}{} algorithm} to optimise for system utility. This algorithm solves the problem of resource allocation in dynamic systems through the use of two sub-algorithms. The \textit{\acronymResourceAllocationAlgorithm{}{}-update algorithm}  learns the resource weightings that maximise the component task values of parent groups' atomic task allocation distributions. The \textit{\acronymResourceAllocationAlgorithm{}{}-weighting algorithm}  combines these into a resource weightings model that is applied across all incoming tasks. The process followed when a task is assigned to a parent agent is shown in Figure \ref{fig:mgrao-highlevel} with the flow of weight modelling and combination shown in Figure \ref{fig:mgrao-overview-compressed}.

The \textit{\acronymResourceAllocationAlgorithm{}{} update} algorithm is applied when a child agent $\varChildAgent{}{}$ receives the absolute task value   $\absoluteTaskValue(\varCompositeTask{}{}, \varAtomicTask{}{})$  from a parent agent $\varParentAgent{}{}$ for completing an atomic task type $\varAtomicTaskType{}{} = \atomicTaskTypeFunction{}{}(\varAtomicTask{}{})$ of the parents' composite task $\varCompositeTask{}{}$. We find the parent-task type index $\functionParentTaskIndex{}{}$, allowing the update of the correct value for the child agents' eligibility trace matrix using equation (\ref{eqn:eligibility_trace_update}). The updated eligibility trace matrix is then multiplied by the value of $\absoluteTaskValue(\varCompositeTask{}{}, \varAtomicTask{}{})$, and a fixed learning rate factor $\alpha \in [0,1]$. The resulting matrix is then added to the current resource weights matrix, $\matrixResourceWeight{}{}$ and sum-normalised row-wise.
\begin{algorithm}[ht]
	\SetAlgoLined
	\DontPrintSemicolon
	{
		\KwIn{$\varAtomicTaskType{}{}$, The task type completed}
		\KwIn{$\varParentAgent{}{}$, The parent agent that allocated the task}	
		\KwIn{$\matrixEligibilityTrace{}{}$, The eligibility trace}		\KwIn{$\matrixResourceWeight{}{}$, The resource allocation weights matrix}
		\KwIn{$\absoluteTaskValue(\varCompositeTask, \varAtomicTask)$, The absolute value for the completed task}
		
		\KwIn{$\varLearningRate{}{}$, The learning rate}
		\KwIn{$\gamma$, The decay rate}
		\;
		\KwOut{$\matrixResourceWeight{}{}$, The updated resource allocation weights matrix}
		\;
		\tcp{Update the eligibility trace}
		$\matrixEligibilityTrace{}{} \funcupdate \functionEligibilityTraceUpdateSignature{}{}$ \;
		\tcp{Apply the eligibility trace to resource weights}
		$\matrixResourceWeight{}{} \funcupdate \matrixResourceWeight{}{} + \varLearningRate{}{} \absoluteTaskValue(\varCompositeTask, \varAtomicTask) \matrixEligibilityTrace{}{}$ \; 
		\tcp{Sum-normalise resource weight matrix rows}
		$\matrixResourceWeight{}{} \leftarrow \functionSumNormSignature{\matrixResourceWeight{}{}}{}$ \;
		\Return $\matrixResourceWeight{}{}$	\;
	}
	\caption{The \acronymResourceAllocationAlgorithmExtended{}{} update algorithm}
	\label{alg:multi_channel_priority_optimisation_by_functional_approximation_update}
\end{algorithm}

The \textit{\acronymResourceAllocationAlgorithm{}{} weighting} algorithm is used when an incoming atomic task is being carried out by a child agent. It uses the combined resource weight equation (\ref{eqn:combined_resource_weights}) to generate the vector of combined resource weights from the resource weights matrix $\matrixResourceWeight{}{}$. The atomic task type selects the index of the resource weight to be applied to the task, $\varResourceWeighting{\varChildAgent{}{}, \varTime{}{}}{} (\varAtomicTaskType{}{})$, then returns the actual resource allocation, $(\varResource{}{}, \varResourceWeighting{\varChildAgent{}{},\varTime{}{}}{} \times \functionResourceMap{}{})$, for the task.

\begin{algorithm}[ht]
	\SetAlgoLined
	\DontPrintSemicolon
	{
		\KwIn{$\varAtomicTaskType{}{}$, The task type being performed}
		\KwIn{$\varChildAgent{}{}$, The child agent that is performing the task}
		\KwIn{$\varParentAgent{}{}$, The parent agent that allocated the task}
		\KwIn{$\varResource{}{}$, The resource requiring allocation}
		\KwIn{$\matrixResourceWeight{}{}$, The resource allocation weights matrix}
		\;
			\KwOut{The resource weight transformed task quality}
		\;
		
		\tcp{Find the index in the resource matrix}
		$(i, j) \funcupdate \functionParentTaskIndex{}{}$ \;
		\tcp{Calculate the combined resource weights}
		$\matrixCombinedResourceWeights{}{} \funcupdate \functionCombinedResourceWeightsSignature{}{}$ \;
		
		\tcp{Retrieve the resource weighting for $\varAtomicTaskType{}{}$}
		$\varResourceWeighting{\varChildAgent{}{},\varTime{}{}}{} \funcupdate \matrixCombinedResourceWeights{}{}_{j}$ \;
		\tcp{Return the resource allocation for tasks of this type}
		\Return $(\varResource{}{}, \varResourceWeighting{\varChildAgent{}{},\varTime{}{}}{} \times \functionResourceMap{}{})$ \;
	}
	\caption{The \acronymResourceAllocationAlgorithmExtended{}{} weighting algorithm}
	\label{alg:multi_channel_priority_optimisation_by_functional_approximation}
\end{algorithm}
\section{Evaluation}
\label{section:analysis}

We simulated four systems to evaluate the \acronymResourceAllocationAlgorithm{}{} algorithms' performance. The \textit{single-child} system evaluated the performance of \acronymResourceAllocationAlgorithm{}{} where a child agents' incoming tasks had a stable distribution, as the parent agents competed over the resource allocation of one agent only. 
In the \textit{multi-child} system parent agents could choose between a number of child agents to allocate atomic tasks to, with a random probability, $\epsilon$, that they would allocate to a non-optimal child agent. Non-optimal child agents were selected based on the \textit{boltzmann distribution}\footnote{The probability of choosing a child agent $\varChildAgent{}{}$ with atomic task quality $\varAtomicTaskQuality{}{}$ from $N$ other agents is, $\functionBoltzmann{\varChildAgent{}{}}{}$} of their absolute task values, as found by the parent agents' previous task allocations. This system tested the \acronymResourceAllocationAlgorithm{}{} algorithm where child agents' parent groups had variable incoming task distributions. To evaluate the effect of dynamism on performance we simulated a \textit{volatile} system where parent-agents had a fixed probability of leaving or re-joining the system each episode. Finally, the \textit{large} system tested the effect of varying parent-agent group size when there were a large number of parent agents allocating tasks to each child agent, to examine the scalability of the algorithm.
\begin{table}[H]
	\captionsetup{labelfont=bf,singlelinecheck=on}
	\caption{Summary of algorithm labels}
	\label{table:summary_of_algorithms}
	\begin{tabular}
		{|p{0.15\textwidth}|p{0.75\textwidth}|}
		\hline
		\textbf{Algorithm} & \textbf{Summary}\\
		\hline
		\simulationUniform{}{} &  Resource allocation weights of child agents are fixed to uniform values. \\
		\simulationMgraoOne{}{} & This system uses only one parent agent group per-child agent rather than multiple groups. This is similar to state-of-the-art distributed Q-learning algorithms.\\
		\simulationMgraoMax{}{} & Every parent agent is placed in its own parent-agent group. \\
		\simulationMgraoXY{}{} & These algorithms have an $x$:$y$ ratio of parent-agent groups to each child-agent. \\
		\hline
	\end{tabular}
\end{table}

Labels for the algorithms and configurations used in the simulations are described in Table \ref{table:summary_of_algorithms}. General system parameters and individual system parameters are shown in Tables \ref{table:general_parameter_values} and \ref{table:simulation_parameter_values} respectively, included in Appendix \ref{section:parameters}. The composite task frequency distribution introduced the same fixed set of tasks over a defined period, giving each \textit{episode} of the system. The parent agent groups for each child agent were fixed throughout the simulation. Each child agents' available resources were set at system start time in the range $(0, 1] \in \setRealNumbersNonNegative{}{}$ drawn randomly from a \textit{normal distribution}\footnote{A normal distribution defined by values in $X \sim \mathcal{N}(\mu,\sigma{^2}), \mu=0.5,\sigma=0.2$}.
\label{section:experimental}
\begin{table}[ht]
\captionsetup{labelfont=bf,singlelinecheck=on}
\caption{Experimental results for single child system after 100 episodes}
\label{table:experimental_results_single_child}
\begin{tabular}[t]
{|p{0.15\textwidth}|p{0.15\textwidth}| p{0.15\textwidth}| p{0.15\textwidth}|}
\hline
\textbf{Algorithm} & \textbf{\% from \simulationUniform{}{}} & \textbf{\% from \simulationMgraoMax{}{}} & \textbf{\% of \simulationMgraoMax{}{}}\\
\hline
\simulationMgraoMax{}{} & $\resultSingleMgraoMax{}{}$ & $-\resultMgraoMaxToMax{}{}$ & $\resultMgraoMaxPercentageMax{}{}$ \\
\simulationMgraoOne{}{} & $\resultSingleMgraoOneOne{}{}$ & $-\resultSingleMgraoMaxToOneOne{}{}$ & $\resultSingleMgraoOneOnePercentageMax{}{}$ \\
\hline
\end{tabular}
\end{table}

\begin{table}[ht]
\captionsetup{labelfont=bf,singlelinecheck=on}
\caption{Experimental results for multi-child system after 100 episodes}
\label{table:experimental_results_multi_child}
\begin{tabular}[t]
{|p{0.15\textwidth}|p{0.15\textwidth}| p{0.15\textwidth}| p{0.15\textwidth}|}
\hline
\textbf{Algorithm} & \textbf{\% from \simulationUniform{}{}} & \textbf{\% from \simulationMgraoMax{}{}} & \textbf{\% of \simulationMgraoMax{}{}}\\
\hline
\simulationMgraoMax{}{} & $\resultMultiMgraoMax{}{}$ & $-\resultMgraoMaxToMax{}{}$ & $\resultMgraoMaxPercentageMax{}{}$ \\
\simulationMgraoOne{}{} & $\resultMultiMgraoOneOne{}{}$ & $-\resultMultiMgraoMaxToOneOne{}{}$ & $\resultMultiMgraoOneOnePercentageMax{}{}$ \\
\hline
\end{tabular}
\end{table}

\begin{table}[ht]
\captionsetup{labelfont=bf,singlelinecheck=on}
\caption{Experimental results for volatile system after 100 episodes}
\label{table:experimental_results_volatile_system}	
\begin{tabular}[t]
{|p{0.15\textwidth}|p{0.15\textwidth}| p{0.15\textwidth}| p{0.15\textwidth}|}
\hline
\textbf{Algorithm} & \textbf{\% from \simulationUniform{}{}} & \textbf{\% from \simulationMgraoMax{}{}} & \textbf{\% of \simulationMgraoMax{}{}}\\
\hline
\simulationMgraoMax{}{} & $\resultVolatileMgraoMax{}{}$  & $-\resultMgraoMaxToMax{}{}$ & $\resultMgraoMaxPercentageMax{}{}$ \\
\simulationMgraoOne{}{} & $\resultVolatileMgraoOneOne{}{}$ & $-\resultVolatileMgraoMaxToOneOne{}{}$ & $\resultVolatileMgraoOneOnePercentageMax{}{}$ \\
\hline

\end{tabular}
\end{table}

\begin{table}[ht]
\captionsetup{labelfont=bf,singlelinecheck=on}
\caption{Experimental results for large system after 100 episodes}
\label{table:experimental_results_large_system}
\begin{tabular}[t]
{|p{0.15\textwidth}|p{0.15\textwidth}| p{0.15\textwidth}| p{0.15\textwidth}|}
\hline
\textbf{Algorithm} & \textbf{\% from \simulationUniform{}{}} & \textbf{\% from \simulationMgraoMax{}{}} & \textbf{\% of \simulationMgraoMax{}{}}\\
\hline
\simulationMgraoOne{}{} & $\resultLargeMgraoOneOne{}{}$ & $-\resultLargeMgraoMaxToOneOne{}{}$ & $\resultLargeMgraoOneOnePercentageMax{}{}$ \\
\simulationMgraoMicro{}{} & $\resultLargeMgraoTwoOne{}{}$ & $-\resultLargeMgraoMaxToTwoOne{}{}$ & $\resultLargeMgraoTwoOnePercentageMax{}{}$ \\
\simulationMgraoTiny{}{} & $\resultLargeMgraoFiveOne{}{}$ & $-\resultLargeMgraoMaxToFiveOne{}{}$ & $\resultLargeMgraoFiveOnePercentageMax{}{}$ \\
\simulationMgraoSmall{}{} & $\resultLargeMgraoTenOne{}{}$ & $-\resultLargeMgraoMaxToTenOne{}{}$ & $\resultLargeMgraoTenOnePercentageMax{}{}$ \\
\simulationMgraoMedium{}{} & $\resultLargeMgraoTwentyFiveOne{}{}$ & $-\resultLargeMgraoMaxToTwentyFiveOne{}{}$ & $\resultLargeMgraoTwentyFiveOnePercentageMax{}{}$ \\
\simulationMgraoMax{}{} & $\resultLargeMgraoMax{}{}$ & $-\resultMgraoMaxToMax{}{}$ & $\resultMgraoMaxPercentageMax{}{}$  \\
\hline
\end{tabular}
\end{table}
Results for the single-child, multi-child, volatile and large systems are shown in Tables \ref{table:experimental_results_single_child}, \ref{table:experimental_results_multi_child}, \ref{table:experimental_results_volatile_system} and \ref{table:experimental_results_large_system}. The percentage system utility improvement for the system using the algorithm specified rather than uniform resource allocation is shown. Percentage utilities are also included for each algorithm in comparison to the \simulationMgraoMax{}{} algorithm for each system.

\subsubsection*{Single child agent performance}
\begin{figure}[ht]
	\centering
	\includegraphics[width=0.8\linewidth]{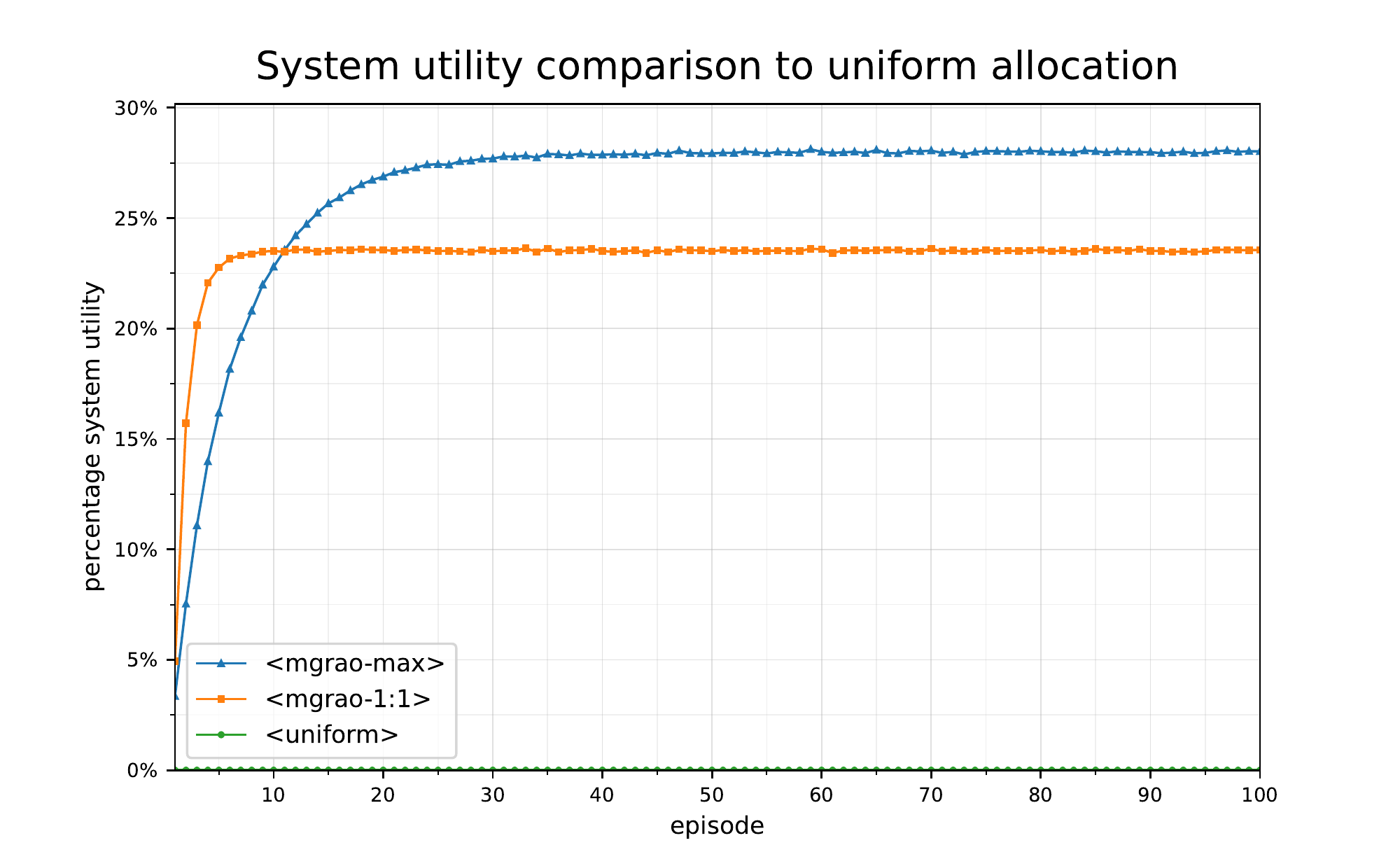}
	\captionsetup{labelfont=bf,singlelinecheck=on}
	\caption{System utility performance comparison to uniform allocation - single child agent}
	\label{fig:resource-allocation_task-allocation-value-utility-comparison}
	\Description[Single child agent results]{How system utility is optimised with a single child agent}
\end{figure}
As seen in Figure \ref{fig:resource-allocation_task-allocation-value-utility-comparison}, \simulationMgraoMax{}{} shows a $\resultSingleMgraoMax{}{}$ improvement in performance as compared to the \simulationUniform{}{} case. It also performs $\resultSingleMgraoMaxToOneOne{}{}$ better than \simulationMgraoOne{}{} after $100$ episodes. Although the \simulationMgraoOne{}{} algorithm initially learns to improve its allocation policy more quickly than \simulationMgraoMax{}{}, by episode $10$ its performance flattens out and \simulationMgraoMax{}{} surpasses it. As well as the performance improvements in allocation optimality, these results demonstrate how the learning of multiple resource allocations for groups of agents before blending allows the algorithm to learn a more complex function approximation. 
\begin{figure}[ht]
	\centering
	\includegraphics[width=0.8\linewidth]{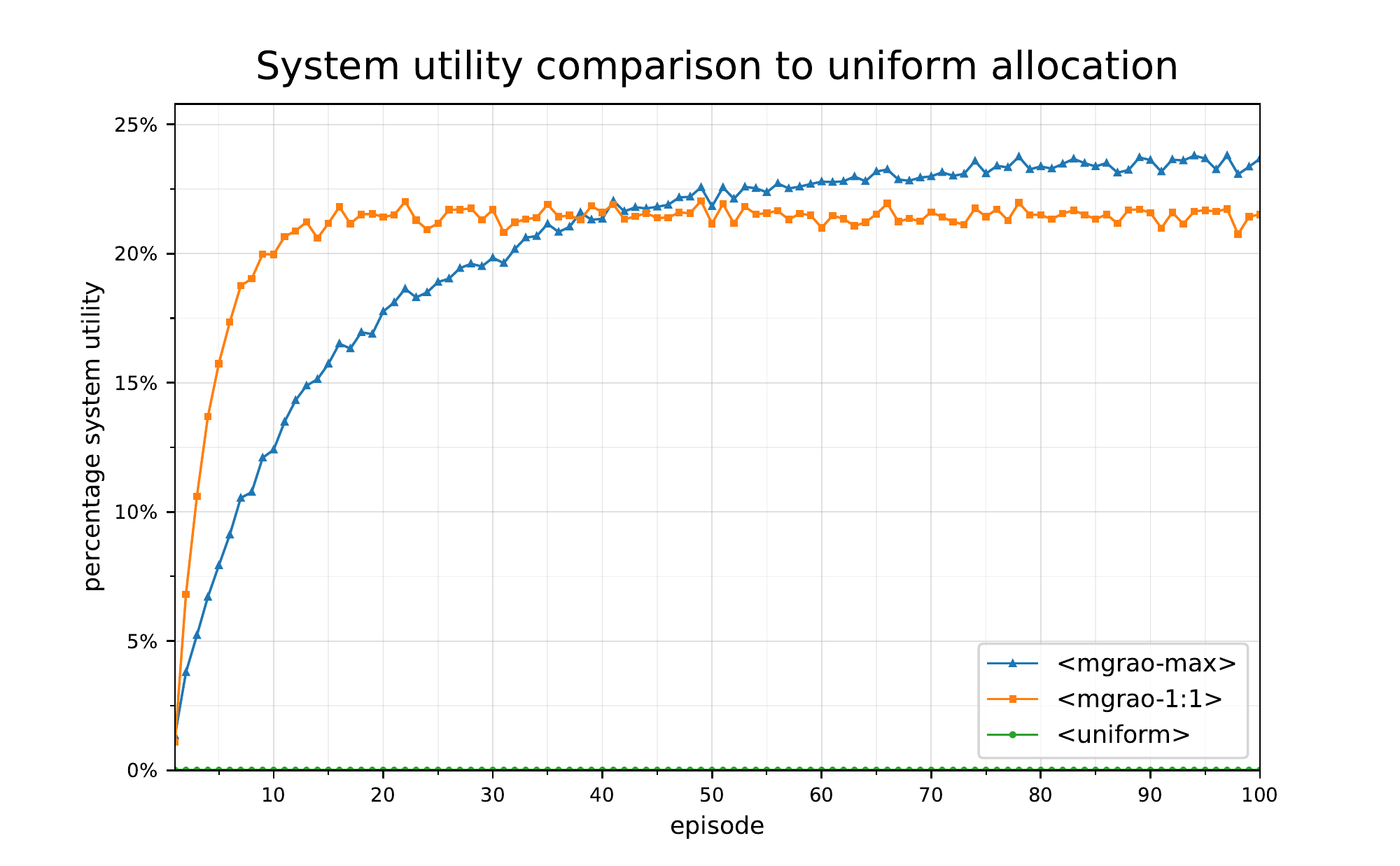}
	\captionsetup{labelfont=bf,singlelinecheck=on}
	\caption{System utility performance comparison to uniform allocation - multiple child agents}
	\label{fig:resource-allocation-multi_task-allocation-value-utility-comparison}
	\Description[Multiple child agent results]{How system utility is optimised with multiple child agents}
\end{figure}
\subsubsection*{Performance with choice of multiple child agents:}
From Figure \ref{fig:resource-allocation-multi_task-allocation-value-utility-comparison} we can see that  \simulationMgraoMax{}{} performs $\resultMultiMgraoMax{}{}$ better  than \simulationUniform{}{} and $\resultMultiMgraoMaxToOneOne{}{}$ better than \simulationMgraoOne{}{} after $100$ episodes. As the \simulationMgraoOne{}{} simulation treats all of a child agents' incoming atomic tasks as one, it does not retain parent agent task allocation patterns through time as well as when multiple parent agent groups are used. Therefore, when a parent agent stops allocating tasks to it, knowledge of the learnt task allocation distribution for that agent is quickly lost. With \simulationMgraoMax{}{}, different optimal resource weighting distributions are learnt for each parent-agent group. When a parent agents task allocations are intermittent, the child agents' knowledge of that agents best-known resource weightings distribution persists for longer. Thus, when a parent agents' task allocation to a child agent is temporarily disrupted through effects such as exploring allocating atomic tasks to other agents, or losing connectivity, the child agent is able to reuse past learnt information when the parent agent start allocating tasks to it again.  
\begin{figure}[ht]
	\centering
	\includegraphics[width=0.8\linewidth]{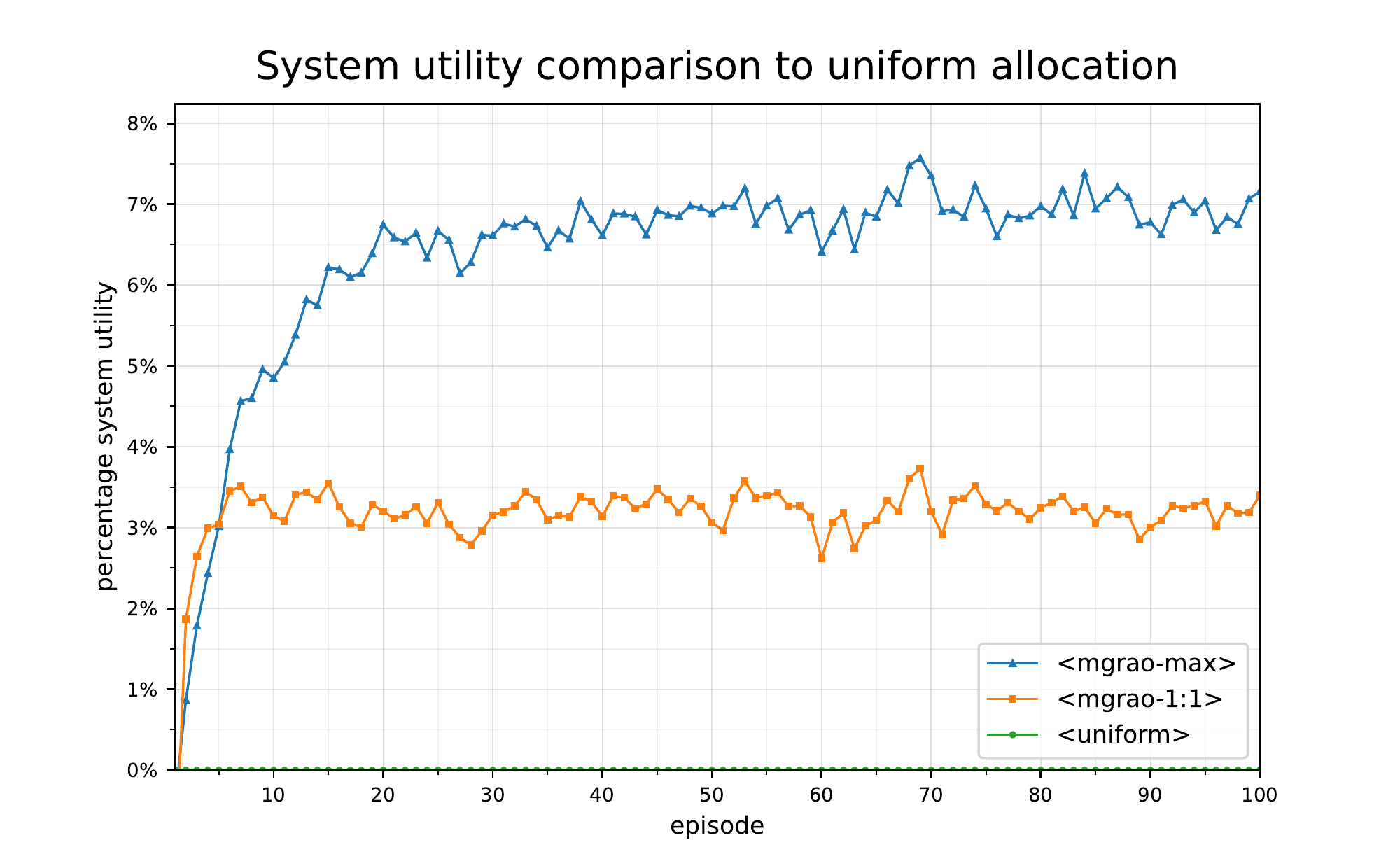}
	\captionsetup{labelfont=bf,singlelinecheck=on}
	\caption{System utility performance comparison to uniform allocation - volatile system}
	\label{fig:resource-allocation-volatile_task-allocation-value-utility-comparison}
	\Description[System utility results in a volatile system]{System utility is optimised within a system with agent volatility}
\end{figure}
\subsubsection*{Performance under agents leaving and joining the system:} 
\begin{figure}[ht]
	\centering
	\includegraphics[width=0.8\linewidth]{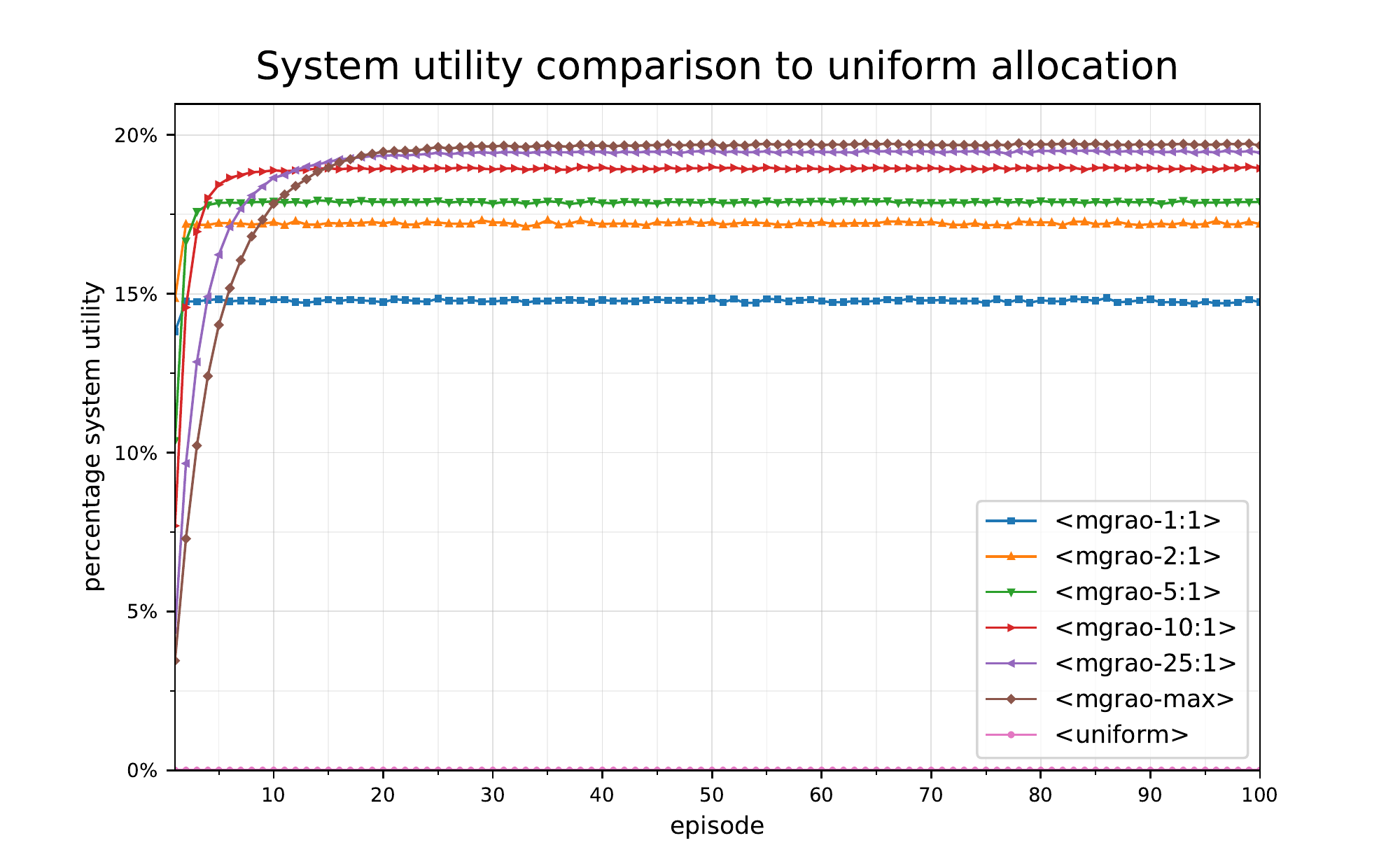}
	\captionsetup{labelfont=bf,singlelinecheck=on}
	\caption{System utility performance comparison to uniform allocation - large system}
	\label{fig:resource-allocation-large_task-allocation-value-utility-comparison}
	\Description[System utility results in as the number of agents increase]{System utility is optimised even as the number of agents in the system grows}
\end{figure}

In the volatile system simulation, shown in Figure \ref{fig:resource-allocation-large_task-allocation-value-utility-comparison}, we see \simulationMgraoMax{}{} perform $\resultVolatileMgraoMax{}{}$ better than \simulationUniform{}{}. In contrast, 
\simulationMgraoOne{}{} only achieves $\resultVolatileMgraoOneOnePercentageMax{}{}$ of the improvement of \simulationMgraoMax{}{} under the same conditions. As parent agents leave and join the system at random, the use of parent-agent groups allows \acronymResourceAllocationAlgorithm{}{} to retain the knowledge the child agent has about these agents' task allocation patterns for a longer period of time than when using fewer parent-agent groups. As agents that had previously left the system rejoin (or regain connectivity), the child agent can re-apply this previous knowledge without having to re-learn the parent agents' task allocation distribution. This means \simulationMgraoMax{}{} will have a higher system utility under volatile conditions than \simulationMgraoOne{}{}, which loses this knowledge more quickly.

\subsubsection*{Performance changes with parent-agent group size in a large system} 
In the large system \simulationMgraoMax{}{} uses the maximum number of parent-agent groups and achieves the best performance, at $\resultLargeMgraoMax{}{}$ over \simulationUniform{}{}. The algorithms \simulationMgraoMedium{}{}, \simulationMgraoSmall{}{}, \simulationMgraoTiny{}{}, \simulationMgraoMicro{}{}, \simulationMgraoOne{}{} achieve
$\resultLargeMgraoTwentyFiveOnePercentageMax{}{}$,
	$\resultLargeMgraoTenOnePercentageMax{}{}$,
$	\resultLargeMgraoFiveOnePercentageMax{}{}$,
$	\resultLargeMgraoTwoOnePercentageMax{}{}$, and
$	\resultLargeMgraoOneOnePercentageMax{}{}$
of the performance of \simulationMgraoMax{}{} respectively. Though \simulationMgraoOne{}{} was still $\resultLargeMgraoOneOne{}{}$ above \simulationUniform{}{} performance. These results demonstrate how the performance of the more scalable, size-constrained parent-agent group algorithms still perform close to the algorithms' optimum, given by \simulationMgraoMax{}{}, where unlimited resources dedicated to learning are assumed. In the \simulationMgraoXY{}{} configurations, \acronymResourceAllocationAlgorithm{}{} optimises its resource usage based on the distribution of task allocations and the per-parent-agent component task values that result from them, rather than requiring individually learnt values for each agent. As such, modelling based on aggregations of these values can perform well, depending on factors such as their similarity, or how stable the values are for individual parent agents.
\section{Conclusions and future work}
\label{section:conclusions}

This work looked at the problem of how agents can distribute their resources across multiple tasks, which have been allocated to them as part of other agents' composite tasks, to improve the utility of the system. The solution proposed uses the \acronymResourceAllocationAlgorithm{}{} algorithm to group together an agents' tasks by the parent agents that allocated them, model the value of these tasks to the parent agents over time, then combine those models to give a resource allocation that will improve the system utility overall. The evaluation results showed that this solution performed $\resultRangeFixedMgrao{}{}$ better than a uniform allocation policy in the simulated system. As a child agent learnt models of the value to the parent agents of the tasks assigned to it, it was able to better distribute its resources to maximise utility. The use of multiple models of incoming tasks split across parent-agent groups allowed a more detailed resource allocation model to be learnt than when only one group was used, and also gave the algorithm robustness under volatility and variations in the distribution of tasks. 

Evaluation was performed in systems ranging from $1-3$ child agents being allocated tasks from between $10-50$ parent agents, representing common ranges found in real-world systems such as in V2X. The addition of parent-agent groups, and their scalable performance, is important to the algorithms' applicability across a large range of system sizes. As they aggregate parent agents into a fixed number of groups, they act as a constraint on the resource overhead required to run the algorithm. This helps the solution be applicable to larger systems.

Future work will look at the applications of the algorithm in real-world environments such as resource distribution in wireless sensor networks, cloud-computing environments, adaptive service-composition, as well as V2X systems. In addition there are areas of improvement for the algorithm itself.

In the case where a child agent does not have availability of a resource needed for a parent agent request, it would return a failure. Work has been done to extend the algorithm to be more robust in completing incoming tasks in this situation. \textit{Hierarchical resource allocation} would mean that a child agent could access resources through other agents in addition to its own, allowing more of its allocated tasks to be completed. As an example, an agent in a cloud-computing environment might be running a computation task for another agent and run out of memory resource. It would then be able to communicate with another agent to utilise some of its available memory, or re-allocate the task to it, to complete the task.

Currently, a parent agent with an unpredictable task allocation distribution would be assigned a parent-agent group randomly, which may contain agents with stable distributions. For example, a parent-agent could be a faulty node in a network, making its task allocations unpredictable compared to fully working nodes. As these distributions are combined for modelling, the more random distribution reduces the child agents' ability to learn a useful model for that parent-agent group. For this reason, adaptive parent-agent groups could be investigated, where the makeup of these groups changes according to the variability of frequency distribution of parent agents' resource requests. This would allow child agents to learn and apply models for more predictable parent-agent groups, and reduce the impact of unpredictable ones on its resource allocation strategy.

\label{section:references}
\bibliographystyle{acm}
\bibliography{bibliography}

\begin{appendix}
\section{Mathematical notation}
\label{section:symbols}
\begin{table}[H]
\captionsetup{labelfont=bf,singlelinecheck=on,justification=raggedright}
\caption{Summary of agent symbols}
\label{table:summary_of_agent_symbols}
\centering
\small
\begin{tabular}{|p{0.1\textwidth}|p{0.8\textwidth}|}
	\hline
$\varSystem{}{}$ & The multi-agent system \\
$\setParentAgent{}{}$ & A set of parent agent   \\
$\setChildAgent{}{}$ & A set of child agent  \\
$\setAtomicTaskType{}{}$ & A set of atomic task types  \\
$\setCompositeTaskType{}{}$ & A set of composite task types   \\
$\varTaskGroupMap{}{}$ & A mapping of composite task types to parent agents that ensure completion of tasks of that type    \\
$\setAtomicTask{}{}$ & A set of atomic tasks   \\
$\varAtomicTaskInstanceDetails{}{}$ & A specification of an atomic task   \\
$\setCompositeTask{}{}$ & A set of composite tasks   \\
$\atomicTaskTypeFunction{}{}$ & A mapping from atomic tasks to task type  \\
$\compositeTaskTypeFunction{}{}$ & A mapping from composite tasks to task type   \\
$\varFrequencyFunction{}{}$ & A mapping of composite tasks to frequency distribution of arrival in the system   \\
$\varTime{}{}$ & Time   \\
$q_{cg}$ & The quality produced by a child agent for a task \\
\hline
\end{tabular}
\end{table}

\begin{table}[H]
\captionsetup{labelfont=bf,singlelinecheck=on,justification=raggedright}
\caption{Summary of resource symbols}
\label{table:summary_of_resource_symbols}
	\centering
	\small
\begin{tabular}{|p{0.1\textwidth}|p{0.8\textwidth}|}
\hline
$ctv$ & The component tasks proportional value, the fractional value of the contribution of its atomic tasks \\
$\setResource{}{}$ & Set of resource needed to perform tasks \\
$\varResourceMap{}{}$ & Resource map \\

$\varTaskAllocation{}{}$ & Task allocation \\

$\varComponentTasksResult{}{}$ & Composite task result,  \\
$\varTaskAllocationQuality_{\varTime}(\varCompositeTask)$ & Composite task quality, the quality of a composite task \\
$\varComponentTasksValue{}{}$ & Component task value \\
$\varResourceWeighting{\varChildAgent{}{}, \varTime}{}$ & Resource weighting, the proportion of a resource an agent devotes to a task type \\
$\varResourceAllocation_{\varChildAgent{}{}}$ & Agent resource allocation, the amount of each resource a child agent devotes to each type type at a given time \\
$\sysResourceAllocation_{\varTime}$ & System resource allocation, the set of all agents' resource allocations in the system at a given time \\
$\varTaskAllocation{}{}$ &Task allocation, a mapping from atomic tasks to the child agents that are allocated to perform them \\

$\absoluteTaskValue{}{}$ & Component tasks absolute value, The absolute value of each component atomic task of a composite task executed at a given time \\
$\varSystemUtility(\setTime)$ & System utility, the utility of a system in a given time period \\
\hline
\end{tabular}
\end{table}

\begin{table}[H]
\captionsetup{labelfont=bf,singlelinecheck=on,justification=raggedright}
\caption{Summary of solution symbols}
\label{table:summary_of_solution_symbols}
	\centering
	\small
\begin{tabular}{|p{0.2\textwidth}|p{0.7\textwidth}|}
\hline
$\functionPAG{}{}$ & Parent agent group, maps a child agents' parents agents into fixed groupings \\
$\functionPGroupSet{}{}$ & The set of all parent agent groups of $\setParentAgent{}{}$ of a child agent $\varChildAgent{}{}$ \\
$\functionPAGResourceWeighting{}{}$ & Parent group weights, the resource weightings associated with a parent group $\setParentAgent{}{}$ of a child agent $\varChildAgent{}{}$ \\
$\matrixResourceWeight{}{}$ & Parent group weights matrix, the parent group weights for all parent groups of a child agent \\
$\functionPAGSampleCount{}{}$ & Parent agent sample count, a mapping of parent agents to a count of tasks allocated to a child agent $\varChildAgent{}{}$ by a member of the group\\
$\functionKullbackLiebler{}{}$ & Parent group weights entropy, the relative entropy of the resource weights of parent group $\setParentAgent{}{} \in \functionPGroup{}{}$ for a child agent $\varChildAgent{}{}$ \\
$\setWeightBlending{}{}$ & Resource weights blending vector, a vector that combines sample counts and resource weight entropy for each parent-group \\
$\functionCombinedResourceWeightsSignature{}{}$ & Combined resource weights function, maps a child agents' parent group weights and blending values to a set of resource weights \\
$\functionParentTaskIndex{}{}$ & Parent-task type index mapping, utility function to map a specific parent  agents and task-type to an index in a child agents' parent agent weights matrix \\
$\matrixEligibilityTrace{}{}$ & Eligibility trace matrix, a matrix of values to be used to map absolute task values to multiple past actions  \\
$\functionEligibilityTraceUpdateSignature{}{}$ & Eligibility trace matrix update, algorithm used to update the eligibility trace matrix \\
\hline
\end{tabular}
\end{table}
\section{Parameters for system simulations and algorithms}
\label{section:parameters}

\begin{table}[ht]
\captionsetup{labelfont=bf,singlelinecheck=on,justification=raggedright}
\caption{General parameter values}
\label{table:general_parameter_values}
\centering
\small
\begin{tabular}{|p{0.1\textwidth}|p{0.6\textwidth}| c|}
\hline
\textbf{Variable} & \textbf{Summary} & \textbf{Value}\\
\hline
$\funcSize{\setAtomicTaskType{}{}}{}$ & Number of atomic task types & $20$ \\
$\funcSize{\setCompositeTaskType{}{}}{}$ & Number of composite task types & {$10$} \\
$\funcSize{\varAtomicTaskType{}{} \in \varCompositeTaskType{}{}}{}$ & Number of atomic tasks composing a composite task type & {$5$} \\
$\varFrequencyFunction{}{}$ & Frequency distribution of composite tasks' arrival in the system& {One $\varCompositeTaskType{}{}$ per $\varParentAgent{}{}$ per episode} \\
$\funcSize{\setResource{}{}}{}$ & Number of resource types needed to perform tasks & {$1$} \\
$\varAtomicTaskQuality{}{}$ & The atomic task quality produced by a child agent for a task.  & {$(0, 1]$} \\
$\varComponentTasksValue{}{}$ & The component tasks value produced by a parent agent for atomic tasks that are part of its composite task. & {$(0, 1]$} \\
\hline
\end{tabular}
\end{table}

\begin{table}[ht]
\captionsetup{labelfont=bf,singlelinecheck=on,justification=raggedright}
\caption{Simulation parameter values}
\label{table:simulation_parameter_values}
	\centering
	\small
\begin{tabular}{|p{0.2\textwidth}|p{0.3\textwidth}|cccc|}
\hline
\textbf{Variable} & \textbf{Summary} & \textbf{Single} & \textbf{Multi} & \textbf{Volatile} & \textbf{Large}\\
\hline
$\funcSize{\setParentAgent{}{}}{}$ & Number of parent agents in the system & $10$ & $10$ & $10$ & $50$\\
$\funcSize{\setChildAgent{}{}}{}$ & Number of child agent in the system & $1$ & $3$ & $1$ & $1$\\
$\funcSize{\functionPAG{\setParentAgent{}{}}{}}{}$ & Parent agent group size & $1$ & $1$ & $1$ & $\lbrace 1,2,5,10,25,50 \rbrace$\\
$P(leave/join|\varParentAgent{}{})$ & Probability of parent-agent leaving or re-joining system per-episode & $0$ & $0$ & $0.25$ & $0$\\
\hline
\end{tabular}
\end{table}

\end{appendix}
\end{document}